\pdfoutput=1

\documentclass[11pt]{article}

\usepackage[preprint]{acl}
\usepackage{multirow}
\usepackage{xcolor}         
\usepackage{enumitem}
\usepackage{makecell}
\usepackage{times}
\usepackage{latexsym}
\usepackage{microtype}
\usepackage{hyperref}
\usepackage{graphicx}
\usepackage{subfigure}
\usepackage{amsmath}
\usepackage{amssymb}
\usepackage{listings}
\usepackage{mathtools}
\usepackage[capitalize,noabbrev]{cleveref}
\usepackage{amsthm}
\usepackage{booktabs} 
\usepackage{pifont}
\usepackage[T1]{fontenc}

\usepackage[utf8]{inputenc}

\usepackage{microtype}

\usepackage{inconsolata}

\usepackage{graphicx}

\title{Data-Copilot: Bridging Billions of Data and Humans with\\ Autonomous Workflow}

\author{Wenqi Zhang$^{1}$, Yongliang Shen$^{1}$, Zeqi Tan$^{1}$, Guiyang Hou$^{1}$,\\
       {\bf Weiming Lu$^{1}$, Yueting Zhuang$^{1}$}\\
  $^1$College of Computer Science and Technology, Zhejiang University\\
  \texttt{\{zhangwenqi, luwm\}@zju.edu.cn}\\
  Project Page: \url{https://github.com/zwq2018/Data-Copilot}} 

\begin{document}
\maketitle
\begin{abstract}
Industries such as finance, meteorology, and energy generate vast amounts of heterogeneous data daily. Efficiently managing, processing, and visualizing such data is labor-intensive and frequently necessitates specialized expertise. Leveraging large language models (LLMs) to develop an automated workflow presents a highly promising solution. However, LLMs are not adept at handling complex numerical computations and table manipulations, and they are further constrained by a limited length context. To bridge this, we propose Data-Copilot, a data analysis agent that autonomously performs data querying, processing, and visualization tailored to diverse human requests. The advancements are twofold: First, it is a \textbf{code-centric agent} that leverages code as an intermediary to process and visualize massive data based on human requests, achieving automated large-scale data analysis. Second, Data-Copilot involves a \textbf{data exploration} phase in advance, which autonomously explores how to design universal and error-free interfaces from data, reducing the error rate in real-time responses. Specifically, It imitates common requests from data sources, abstracts them into universal interfaces (code modules), optimizes their functionality, and validates effectiveness. For real-time requests, Data-Copilot invokes these interfaces to address user intent. Compared to generating code from scratch, invoking these pre-designed and well-validated interfaces can significantly reduce errors during real-time requests. We open-sourced Data-Copilot with massive Chinese financial data, such as stocks, funds, and news. Quantitative evaluations indicate that our \text{exploration-deployment} strategy addresses human requests \textbf{more accurate} and \textbf{efficiently}, with good interpretability.
\end{abstract}

\begin{figure*}[h] 
    \centering
    \includegraphics[width=1\linewidth]{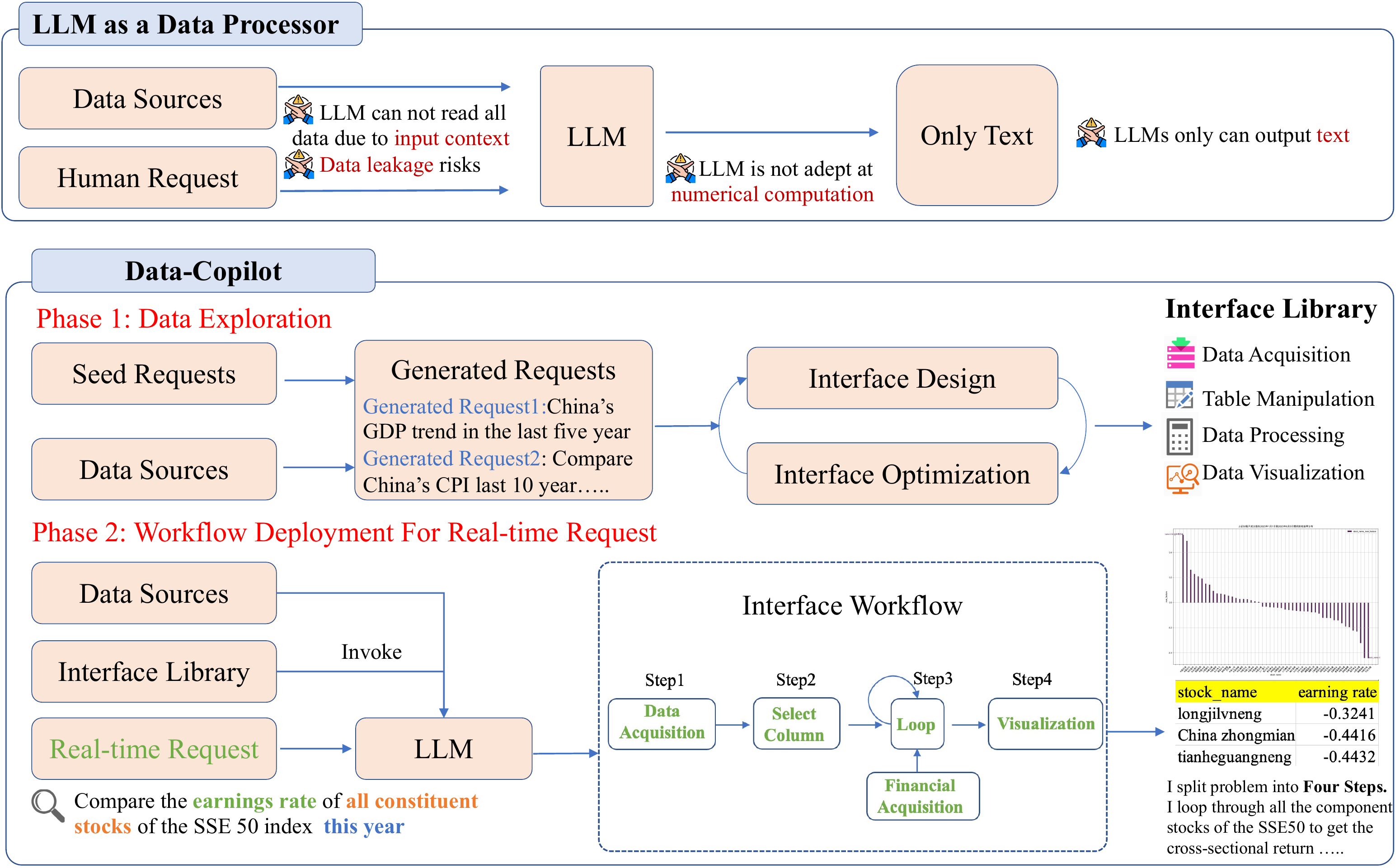}
    \caption{We compare two LLM strategies for automated data analysis. Upper: LLM's capabilities, context length, and output format constraints limit the use of LLMs to process massive data. Bottom: Data-Copilot is a code-centric agent that utilizes code to handle extensive data analysis tasks. It explores how to design more universal and error-free interface modules, improving the success rate of real-time requests. Faced with real-time requests, Data-Copilot invokes self-design interfaces and constructs a workflow for human intent.}
    \label{fig:1}
     \vspace{-5pt}
\end{figure*}

\section{Introduction}
In the real world, vast amounts of heterogeneous data are generated every day across various industries, including finance, meteorology, and energy, among others. Humans have an inherent and significant demand for data analysis because these wide and diverse data contain insights that can be applied to numerous applications, from predicting financial trends to monitoring energy consumption. However, these data-related tasks often require tedious manual labor and specialized knowledge.

Recently, the advancement of large language models (LLMs) \citep{Zeng2023GLM, touvron2023llama, Chatgpt} and techniques~\citep{wei2022chain, kojima2022large} have demonstrated the capability to handle complex tasks. Given the vast amounts of data generated daily, \textbf{can we leverage LLMs to create an automated data processing workflow performing data analysis and visualization} that best matches user expectations?


An intuitive solution is to treat data as a special text, i.e., directly use LLMs to read and process massive data~\citep{wu2023bloomberggpt, zha2023tablegpt}. However, as shown in~\Cref{fig:1}, several challenges must be considered: (1) Due to context limitations of LLMs, it is challenging for LLMs to directly read and process massive data as they do with text. Besides, it also poses the potential risk of data leakage when LLMs directly access private data sources. (2) Data processing is complex, involving many tedious numerical calculations and intricate table manipulations. LLMs are not adept at performing these tasks. (3) Data analysis typically requires visualizing the results of data processing, whereas LLMs are limited to generating text output. These challenges constrain the application of LLMs in data-related tasks.

Recently, many agent-based designs have explored alternative solutions~\citep{wu2023visual,huang2023audiogpt,chen2023agentverse,hong2023metagpt,wu2023autogen,nejjar2023llms,li2024can}. LiDA~\citep{dibia2023lida} and GPT4-Analyst~\citep{cheng2023gpt} focus on exploring insight from data. Sheet-Copilot~\citep{li2023sheetcopilot}, BIRD~\citep{li2023can}, DS-Agent~\citep{DS-Agent}, DB-GPT~\citep{xue2023dbgpt} and TAG~\citep{biswal2024text2sql} apply LLMs to data science domain like Text2SQL. Data Interpreter~\citep{hong2024data} proposes a plan-code-verify paradigm for automating machine learning tasks. These methods showcase the potential of LLMs in completing complex daily tasks through agent design paradigms.



Inspired by this, we advocate leveraging the coding capabilities of LLM to build a data analysis agent. Acting like a human data analyst, it receives human requests and generates code as an intermediary to process massive data and visualize its results (e.g., chart, table, text) for humans. However, creating a code agent that can be used in real-world data analysis tasks is far from an easy feat. \ding{172} LLMs struggle to generate high-quality, error-free code in a single attempt, often containing format errors, logical inconsistencies, or fabricating non-existent functions. \ding{173} Although the inference speed of LLMs has significantly improved, generating lengthy code still consumes a considerable amount of time and tokens. These two challenges—\textbf{high code error rate} and \textbf{inefficient inference}—must be addressed urgently.  




To address this, we observe most human requests are either similar or inherently related. By abstracting common demands into interfaces and validating their functionality in advance, we can significantly improve both the success rate and efficiency of real-time deployment. Therefore, we propose \textbf{Data-Copilot}, an LLM-based agent with an innovative exploration phase to achieve more reliable data analysis. First, Data-Copilot is \textbf{a code-centric agent} that connects massive data sources and generates code to retrieve, process, and visualize data in a way that best matches user’s intent. The code-centric design empowers it to efficiently and securely handle extremely large-scale data and nearly all types of data analysis tasks. Besides, Data-Copilot also incorporates \textbf{a data exploration and interface design phase}. It autonomously explores how to design more universal and error-free interfaces (code modules) based on data schemas in advance. In real-world deployment, Data-Copilot flexibly invokes pre-design interface modules for most requests, deploying a well-verified interface workflow for data processing and visualization. 

Data-Copilot brings three advantages. Firstly, this exploratory process allows Data-Copilot to analyze and summarize the inherent connections between human requests, design general interfaces for similar requests, and pre-validate their correctness, \textbf{reducing errors in real-time responses}. Secondly, when faced with massive requests, our agent only needs to invoke these pre-designed interfaces rather than generate redundant code, significantly \textbf{improving inference efficiency}. Lastly, compared to lengthy code, these interfaces provide \textbf{greater interpretability}, since it is easier for human reading and interaction. To achieve this, we contains three steps (\Cref{fig:1}) when self exploration:


\textbf{Explore data and Synthesize requests}: Data-Copilot discovers potential requests and a broader range of human need from data. It involves a "self-exploration" process to generate massive requests based on all data schemas and seed requests.  

\textbf{Interface Design and Test}: It designs modular interface from synthesized requests, with test cases automatically generated for verification.  

\textbf{Interface Optimization}: To improve versatility, it merges similar interfaces and also revises erroneous interfaces using compiler's feedback.

After \textbf{exploration-design-optimization}, Data-Copilot designs many general and error-free interfaces, e.g., data acquiring, forecasting, and visualizing modules, to accomplish data analysis tasks. When faced with real-time requests, Data-Copilot invokes these predefined interface modules to create a concise interface workflow for user requests. For different requests, Data-Copilot can flexibly deploy various invocation structures, such as step-by-step serial workflows, parallel, or loop. It can even output a hybrid form of interface workflows and raw code for these "unfamiliar" requests. Our contributions are threefold:






\begin{figure*}[!t]
    \centering
    \includegraphics[width=1\linewidth]{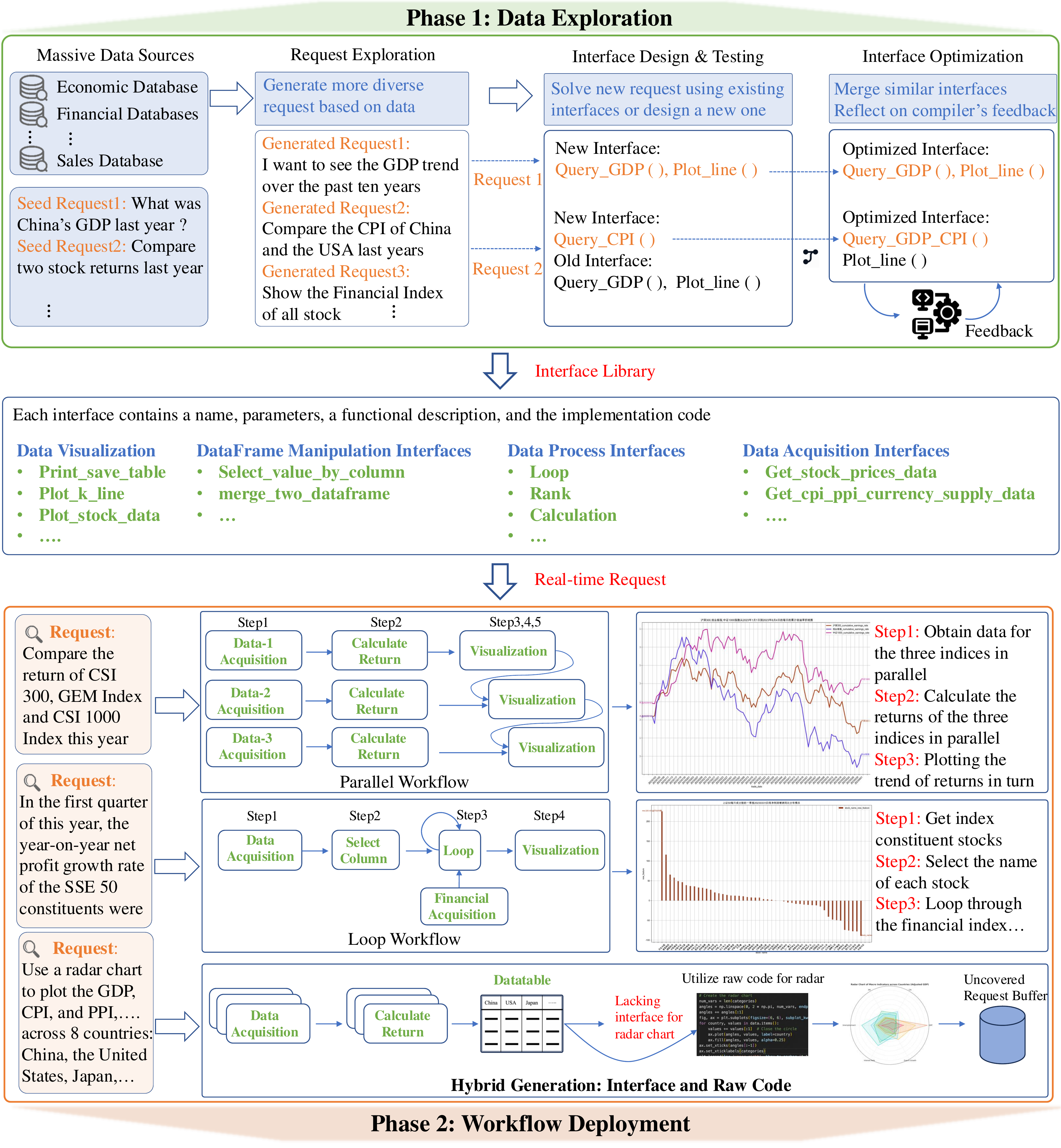}
    \caption{Overview of Data-Copilot. \textbf{Data Exploration:} First, it performs a self-exploration process to uncover potential human requests from data sources. Then it abstracts many universal and error-free interfaces from synthesized requests, including interface designing, testing, and optimizing similar interfaces. This exploration-design-optimization process is operated in advance. \textbf{Workflow Deployment:} Upon receiving real-time requests, Data-Copilot invokes existing interfaces and deploys a workflow for familiar requests, or flexibly combines interfaces and raw code for "uncovered" requests.}
    \label{fig:2}
\end{figure*}

\begin{itemize}
    \setlength{\itemsep}{0pt}
    \item We propose a code-centric agent, Data-Copilot, for automated data analysis and visualization. It leverages LLM's code generation abilities for data querying, processing, and visualization, reducing tedious human labor.
    


    \item We decouple code generation into two phases: data exploration and workflow deployment. In exploration phase, Data-Copilot learns to design universal, error-free interface modules tailored to data. In the face of real-time requests, it flexibly invokes these interfaces to address users' diverse requests. This design enhances the success rate and efficiency of real-time responses.

    \item We open-source Data-Copilot for Chinese financial data analysis, including stocks, funds, and live news. Quantitative evaluations indicate our agent outperforms other strategies, with \textbf{higher success rates and lower token consumption}. Besides, interface workflows are more convenient for human inspection and interaction, offering \textbf{interpretability}.

    
\end{itemize}

\section{Data-Copilot}



Data-Copilot is a code-centric agent capable of performing data analysis and visualization based on human instructions. It operates in two phases: Data Exploration and Workflow Deployment. In the first phase (\cref{section 2.1}), Data-Copilot designs a self-exploration process to discover numerous potential needs based on data schemas (Step 1: exploration). Then based on these synthesized requests, it abstracts many universal interface modules (Step 2: design). After that, it optimizes similar interfaces and tests their correctness, ensuring each interface is correct and universal (Step 3: testing and optimization). The whole process is operated in advance, yielding many generic, error-free interfaces for subsequent use. 

When faced with real-time requests (\cref{section 2.2}), Data-Copilot can flexibly invoke pre-designed interfaces or directly generate raw code based for the user's request. In most cases, existing interfaces can cover the majority of real-world requests, significantly enhancing both success rate and response speed. We provide a detailed prompt for two phases in \Cref{prompt for design} and \ref{prompt for Dispatch}.




\subsection{Data Exploration} \label{section 2.1}
Let's review how human data analysts operate. Initially, they need to observe the available data, understand the data formats, and learn how to access them. Subsequently, humans often design generic modules to simplify the code logic and test these modules for usability. Similar to this, Data-Copilot also autonomously explores data and derives insights from vast data sources, including the relations between the data, and the potential requests associated with the data. Then Data-Copilot abstracts these exploratory insights into numerous reusable code components (interfaces), testing their correctness and optimizing their generality. This process of exploring data, identifying common requests, designing general interfaces, and testing and optimizing their performance is conducted in advance on its own.


\textbf{Self-Exploration Request.}
To explore data and mine insights, we design a self-exploration process. Beginning with some seed requests collected from humans, LLMs are prompted to read the data and generate a large number of requests, each representing a potential demand scenario. This process is similar to~\citep{wang2022selfinstruct, dibia2023lida}, but the LLMs should generate requests specifically based on provided data. As shown in Figure~\ref{fig:2}, when the LLMs observe that the economic database contains historical GDP and CPI data, they generate multiple related requests, e.g., \texttt{Compare the CPI of China and the USA...}, or \texttt{I want to see the GDP trend over the past ten years}. Each request involves one or more types of data.


To achieve this, we first generate a parsing file for each data source to help LLM understand the data. Each file includes a description of the data, the access method, the data schema (the name of each column), and a usage example. Then we feed the parsing files and a few seed requests into the LLMs, and prompt LLMs to synthesize more diverse requests based on these data. The brief example is shown in~\Cref{example for data exploration}.

\textbf{The quality of the generated requests.} A common issue is that the LLM often proposes a request about the data that doesn't exist. To address this, we design a backward verification strategy to check these synthesized requests. Specifically, we instruct the LLM to reverse-convert the generated request into the desired data source and other key information, and then we verify the existence of such data, thereby filtering out hallucinatory requests. Besides, when synthesizing, we also use keywords to control the topics of synthesized requests, ensuring they closely align with real-world distribution. Upon manual evaluation of the synthesized requests, we found that the generated requests generally met our expectations. We discuss the quality of synthesized requests in detail in~\Cref{dataset_statics,human evaluation on seed set and testset}.

\textbf{Interface Design.} After generating massive requests through self-exploration, Data-Copilot abstracts many universal interfaces from these requests. First, we have to clarify what an interface is in our paper. Similar to human-defined functions, an interface is a code module consisting of a name, parameters, a functional description, and an implementation code. It performs specific tasks such as data retrieval, computation, and visualization. Each interface is designed, tested, and optimized iteratively by Data-Copilot.



First, starting from the initial request, we iteratively feed synthesized requests from the self-exploration stage and related data parsing files into LLMs, prompting the LLM to design complete code modules for request solving. Each code module is defined as an interface and stored in the interface library. During each iteration, LLMs are instructed to prioritize utilizing the existing interfaces within the library. If the available interfaces are insufficient for request solving, the LLMs design a new interface.

As shown in Figure~\ref{fig:2}, for the first request: \texttt{I want to see the GDP trend over ...}, Data-Copilot design two interfaces: \texttt{Query-GDP()} and \texttt{Plot-Line()}. As for the second request: \texttt{Compare the CPI of China and the USA over..}, since the previous two could not solve this request, a new interface, \texttt{Query-CPI()}, is designed by LLMs.

\textbf{Interface Testing.} After designing a new interface for a request ($request_i$$\rightarrow$$interface_i$), Data-Copilot autonomously tests its correctness based on compiler feedback. First, Data-Copilot uses the $request_i$ as the seed to generate massive similar requests as test cases. Then it tests the new $interface_i$ one by one. In~\cref{fig:2}, for new $interface_i$: \texttt{Query-GDP()}, Data-Copilot mimics $request_i$ to generate many similar requests to test the \texttt{Query-GDP()}: \texttt{"I want to see USA's GDP over the past 10 years", "I want to see China's GDP for the last year",.}. If the interface passes all test cases, it is retained; otherwise, Data-Copilot self-reflects on error feedback of compilers, correcting the erroneous code snippets until it successfully passes the tests.




\textbf{Interface Optimization.} To optimize the generality of designed interfaces, Data-Copilot also merges similar interfaces. Similar to human developers, each time a new interface is designed, this optimization process is triggered: evaluating whether the newly designed interface can be merged with previous ones.

\begin{itemize} 
    \setlength{\itemsep}{0pt}
    \item \textbf{Retrieve similar interfaces}. After a new interface is designed, we retrieve \textit{Top-N} interfaces from existing interface library. Specifically, we use gte-Qwen1.5-7B-instruct~\citep{li2023towards} to obtain embeddings of each interface code and then calculate their similarity, identifying top N similar interfaces.
    \item \textbf{Decide whether to merge}. LLMs are prompted to compare new interface with \textit{Top-N} retrieved interfaces in terms of functionality, parameters, and processing logic, autonomously deciding whether to merge it with the existing ones. If LLM deems no merging necessary, new interface is retained. As shown in Figure~\ref{fig:2}, two interfaces \texttt{Query-GDP()} and \texttt{Query-CPI()} are merged into \texttt{Query-GDP-CPI()}. This process makes each interface more general and unique.
    \item \textbf{Test the optimized interface}. When two interfaces are merged, Data-Copilot also needs to test the newly merged interface. Test cases from two original interfaces are used to validate the merged interface. The output of the new interface must be consistent with two original interfaces in both format and content.

\end{itemize}


In this phase, Data-Copilot alternates the above three steps: interface design, testing, and optimization until all the requests can be covered by these interfaces. As shown in Figure~\ref{fig:interface}, Data-Copilot designs many interfaces for different task and operation types, e.g., data acquisition, prediction, visualization, and DataFrame manipulation. We provide detailed algorithm processes (\Cref{design_Algorithm}) and examples in \Cref{case interface design}. 

\subsection{Workflow Deployment} \label{section 2.2}

As Figure~\ref{fig:2} shows, it accurately and efficiently handles requests by invoking relevant interfaces (interface workflow). For long-tail or uncovered requests, it flexibly generates raw code (interface-code hybrid strategy) while documenting these for future interface library updates.

\textbf{Interfaces Retrieval.} Considering the significant differences between the interfaces involved in different requests, it is unnecessary to load all interfaces every time. We design a simple yet efficient interface retrieval strategy: hierarchical retrieval. Specifically, we organize all designed interfaces in a hierarchical structure. Each interface is grouped into different tasks (stock task, fund task, etc.) and different operation types (data acquisition, processing, visualization, etc.). Upon receiving a real-time request, Data-Copilot first determines the appropriate task types and required operation types and then loads the interfaces associated with these types for subsequent workflow planning. This can reduce the number of interfaces in the prompt.


\textbf{Interface Invocation Workflow.} After reading interface descriptions, Data-Copilot plans workflows—multiple interfaces in specific order forming chain, parallel, or loop structures. It determines which interfaces to invoke, their sequence, and parameters, outputting in \texttt{JSON} format. Prompts and cases appear in \Cref{prompt for Dispatch,case interface dispatch}.


As Figure~\ref{fig:2} shows, Data-Copilot designs sequential, parallel, or loop interface workflows. For request \texttt{"Compare the return of CSI-300, GEM and CSI-1000 this year"}, it plans a parallel workflow: \texttt{Data Acquisition(), Calculate Return()} for three indices simultaneously, then \texttt{Visualization()}. The second case implements a loop workflow using \texttt{Loop()}.

\textbf{Interface-Code Hybrid Generation.} In the real world, it is inevitable to encounter "uncovered" requests that cannot be addressed by existing interfaces. As a remedy, Data-Copilot adopts an interface-code hybrid generation strategy. Specifically, it prioritizes invoking existing interfaces to resolve user requests. It can solve most common user requests. However, if the deployed workflow continues to fail, or if Data-Copilot proactively determines that current request cannot be resolved by existing interfaces, it would generate raw code directly or generate a combination of raw code and interfaces. This design endows Data-Copilot with the flexibility to handle diverse requests.

Besides, each time Data-Copilot encounters such "uncovered" requests, it also records them in the document. After accumulating sufficient new requests, we re-initiated the Data Exploration and interface design stage (\cref{section 2.1}), i.e., in real-world interactions, we periodically develop new interfaces for emerging demands, continuously updating Data-Copilot's interface library.

\begin{table*}[h!]
    \centering
    \small
    \setlength\tabcolsep{1pt} 
    \centering
    \begin{tabular}{l l c l c c}
    \toprule[1pt]
    \textbf{Name} & \textbf{\makecell[l]{Source}}  & \textbf{\makecell[c]{\#Cases}} & \textbf{Form} & \textbf{Four Types Ratio}  & \textbf{Three Complexity Levels Ratio}  \\
    \midrule
    \makecell[l]{Seed Set} & \makecell[l]{Human-proposed} & 173 & Query &  8.0: 4.0: 4.0: 1 & 2.0: 1.5: 1\\
    \makecell[l]{Set for Interface Design} & \makecell[l]{Self-Exploration} & 3000 & Query &  8.7: 3.3: 4.2: 1 & 1.9: 1.5: 1\\
    Test Set & \makecell[l]{Self-Exploration \& Human} & 547 & Query, Label &   8.6: 3.3: 4.3: 1 & 1.8: 1.4: 1\\
    \bottomrule[1pt]
    \end{tabular}
    \caption{Statistics on human-proposed requests and self-exploration dataset. We report the number of each task type and complexity level. The results indicate that our synthesized requests closely align with human distribution.}\label{dataset_statics}
    \hspace{0cm}
    \vspace{-0.2cm}
\end{table*}

\section{Dataset Synthesis}
\subsection{Environment and Data Sources}
Data-Copilot is developed on Chinese financial market data, encompassing massive stocks, funds, economic data, real-time news, and company financial data. Similar to many works, Data-Copilot utilizes data interfaces provided by Tushare\footnote{\url{https://tushare.pro/}} to access vast amounts of financial data, including time-series data spanning over 20 years for more than 4,000 stocks, funds, and futures. In the first phase, we use a strong LLM (e.g., \texttt{gpt-4o}) for data exploration and a lightweight LLM (e.g., \texttt{gpt-3.5-turbo}) for workflow deployment.

\subsection{The Creation of Dataset} \label{dataset}
\textbf{Human-proposed Request as Seed Set.} We invite 10 students in economics to submit 50 requests each. These requests cover a wide range of common needs, including stock, fund tasks with different complexity levels. These human-proposed requests can represent a distribution of real-world scenarios, where the ratio of four tasks is: Stock(8), Fund(4), Corporation(4) and Others(1). Then we filter out highly similar requests. Lastly, we retain 173 high-quality requests, which are used as seeds for Data Exploration phase.  

\textbf{Self-Exploration Request Set.} As mentioned in~\Cref{section 2.1}, we employ self-exploration to expand the request set. We first feed 173 human-proposed seed requests along with all data descriptions and schema into GPT-4 for data exploration and request synthesis. We use keywords to control the distribution of the synthesized requests, ensuring close alignment with the real world. When exploration, it generates 6480 requests. Then we filter out highly similar requests and retain 3547 requests. Then we adopt a stratified sampling strategy to sample 547 instances as test set from each type. The remaining 3000 requests are used for interface design. The distributions of four task types (stock, fund, corporation, and others) and three complexity levels (single-entity, multi-entities, and multi-entities with complex relations) are shown in~\Cref{dataset_statics}. It shows that synthesized requests closely align with humans. The detailed statistics are shown in \Cref{dataset_statics,fig:dataset_pie,four tasks and three complexities}.

\begin{table*}[t!] 
    \setlength\tabcolsep{3pt} 
    \centering 
    \small
    \begin{tabular}{l|ccccc}  
        \toprule[1pt]
        Model & GPT-4o & GPT-4-turbo & GPT-3.5-turbo & Llama3-70B-Instruct & Llama2-70B-Instruct  \\ 
        \hline
        {Direct-Code} & 73.6 & 70.0 & 28.5 & 52.2 & 29.6 \\ 
        \textbf{Ours} & \textbf{77.9} \scriptsize$\uparrow$4.3 & \textbf{74.6} \scriptsize$\uparrow$4.6 & \textbf{70.2} \scriptsize$\uparrow$41.7 & \textbf{70.9} \scriptsize$\uparrow$18.7 & \textbf{49.3} \scriptsize$\uparrow$19.7  \\
        \midrule
        Model & Codellama-13B & Vicuna-13B-v1.5 & DeepSeek Coder-V2 & Qwen2.5-Coder-32B & Qwen2.5-Coder-14B  \\ 
        \hline
        {Direct-Code} & 21.0 & 17.3 & 63.8 & 66.5 & 51.7  \\ 
        \textbf{Ours} & \textbf{50.8} \scriptsize$\uparrow$29.2 & \textbf{33.2} \scriptsize$\uparrow$15.9 & \textbf{75.3} \scriptsize$\uparrow$11.5 & \textbf{77.2} \scriptsize$\uparrow$10.7 & \textbf{72.3} \scriptsize$\uparrow$20.6  \\        
    \bottomrule[1pt]
    \end{tabular}
    \caption{\texttt{Accuracy} of ours and direct code generation (Direct-Code) using different LLMs for workflow deployment.}
    \hspace{0cm}
    \vspace{-0.2cm}
    \label{tab:data-copilot vs code generation}
\end{table*}
\textbf{Annotate Answer Table for Test Set.}\label{Answer Annotation}
Human annotators are instructed to annotate data tables as answers for 547 test requests. Specifically, annotators manually retrieve and process the corresponding data based on testing request, and record the final data table before chart plotting. For example, request: \texttt{"I want to see China's GDP over past 5 years"}, the labeled data-table is \emph{[2023: 17.8 trillion, 2022: 17.9 trillion, 2021: 17.8 trillion, 2020: 14.6 trillion, 2019: 14.3 trillion]}. 

\textbf{The quality of self-exploration requests.} We manually evaluate the quality of testset. As shown in~\Cref{quality of request}, these synthesized requests show a comparable quality to human-proposed requests.

\subsection{Qualitative Evaluation}
As shown in Figure~\ref{fig:3}, Data-Copilot constructs the invocation workflow step-by-step (each step corresponds to one or more interfaces) and final results (bar, chart, text) for input request.





\textbf{Data-Copilot designs versatile interfaces via data exploration.} After analyzing Chinese financial data and 3,000 self-exploration requests, Data-Copilot created 73 interfaces across five functionalities (data acquisition, index calculation, table manipulation, visualization, general processing). Figure~\ref{fig:interface} shows key interfaces.

\textbf{Data-Copilot adapts to new requests by combining interfaces and raw code.} As in ~\Cref{section 2.2}, for unsupported requests, Data-Copilot integrates existing interfaces with raw code. In ~\Cref{fig:2} (3rd example), it retrieves data via interface, then generates radar chart code, enhancing flexibility for evolving demands.



\textbf{Interface Workflow enhances interpretability.} After execution, Data-Copilot generates visuals and workflow summaries. As shown in \Cref{fig:2,fig:fig_case_pred,fig:fig_case_rank,fig:fig_case_fund,fig:fig_case_stock,fig:fig_case_kline}, outputs are intuitive. Structuring complex code into step-by-step interface calls improves clarity and inspection ease.

\section{Quantitative Evaluation}  
\subsection{Experiments Settings}\label{Experiments Settings} 
\textbf{Baselines and Experiments Details} We compare Data-Copilot with direct code generation (Direct-Code),  ReAct, Reflexion, Multi-agent methods. Detailed prompts are in \Cref{Baselines details,prompt baselines}.

\textbf{Evaluation Methods.} We evaluate all methods on 547 test cases, focusing on three aspects: data-table accuracy, image quality, and inference efficiency. Evaluations are conducted using GPT-4o: \textbf{Data-Table Evaluation}: GPT-4o/turbo compares the generated dataframe with a human-labeled answer table (\Cref{Answer Annotation}). \textbf{Image Evaluation}: To evaluate whether the final image meets human requests, we feed the human request, human-annotated answer table, and the generated image into \texttt{GPT-4o}. \texttt{GPT-4o} is instructed to check image's visual elements based on human request, including numerical points, lines, axes, image aesthetics, and style, i.e., we design a checklist containing 5 categories with 10 sub-dimensions for GPT-4o scoring. The results of data-table and image evaluation are combined as \texttt{Accuracy}. \textbf{Efficiency}: Measured by the total token consumption (\texttt{\#Token}) per method, representing solving efficiency. The evaluation details and comparison between manual evaluation and model evaluation are shown in~\Cref{detail of GPT-4o Evaluation}.


\begin{table}[t!]
    \centering
    \small
    \setlength\tabcolsep{6pt} 
    \begin{tabular}{lcc}
        \toprule[1pt]
        \textbf{Methods}  & \textbf{Accuracy(\%)} & \textbf{\#Token} \\
        \midrule
        Direct-Code            & 28.5 \scriptsize $\pm$2.1 \  & 823  \\ 
        ReAct~\citep{yao2022react}              & 44.1 \scriptsize$\pm$3.4   & 1515 \\ 
        Reflexion~\citep{shinn2023reflexion}          & 59.0 \scriptsize$\pm$3.9  & 2463 \\ 
        Multi-Agent~\citep{hong2023metagpt}        & 57.4 \scriptsize$\pm$1.8  & 2835 \\ 
        Data-Copilot       & \textbf{70.2} \scriptsize$\pm$2.5& \textbf{561.2} \\ \hline
        Data-Copilot + ReAct & 71.5 \scriptsize$\pm$1.7 & 834 \\
        Data-Copilot + Reflexion & 71.8 \scriptsize$\pm$2.2 & 978 \\
        \bottomrule[1pt]
    \end{tabular}
    \caption{Accuracy and efficiency on \texttt{gpt-3.5-turbo}.}
    \hspace{0cm}
    \vspace{-0.2cm}
    \label{main_result_multi_agent}
\end{table}

\begin{table}[t!]
    \centering
    \small
    \setlength\tabcolsep{2pt} 
    
    \begin{tabular}{lcccc}
        \toprule[1pt]
        \textbf{Methods} & \textbf{Single} & \textbf{Multiple} & \textbf{Complex Rel.} & \textbf{Overall} \\
        \midrule
        Direct-Code            & 42.4 & 29.0 & 1.6 & 28.5 \\  
        ReAct        & 56.3 & 48.6 & 14.6 & 44.1 \\
        Reflexion    & 67.8 & 55.1 & 48.1 & 59.0 \\
        Multi-agent  & 63.2 & 64.7 & 35.5 & 57.4 \\  
        Data-Copilot & \textbf{71.8} \scriptsize$\uparrow$4 & \textbf{70.0} \scriptsize$\uparrow$5.3  & \textbf{67.1} \scriptsize$\uparrow$19  & \textbf{70.2} \scriptsize$\uparrow$11.2  \\
        \bottomrule[1pt]
    \end{tabular}
    \caption{\texttt{Accuracy} for three complexity levels samples.}
    \hspace{0cm}
    \vspace{-0.5cm}
    
    \label{table:structure}
\end{table}

\subsection{Comparison Results}

\textbf{Data-Copilot Significantly Reduces Deployment Failure Risk.} As shown in \Cref{tab:data-copilot vs code generation}, compared to direct code generation, Data-Copilot achieves significant improvements: +41.7 (\texttt{GPT-3.5-turbo}), +18.7 (\texttt{Llama3-70B}), and +29.2 (\texttt{Codellama-13B}). By decoupling into interface design and workflow deployment stages, smaller LLMs (\texttt{Llama3-70B}, \texttt{GPT-3.5}) perform at \texttt{GPT-4} level during deployment. Even \texttt{GPT-4} improves by +4\%, confirming generalizability. Pre-designed interfaces reduce errors through optimization and validation during design, whereas direct code generation often overlooks details and introduces mistakes.

\textbf{Data-Copilot Outperforms Advanced Agent Strategies in Both Accuracy and Efficiency.} As shown in \Cref{main_result_multi_agent}, Data-Copilot surpasses all baseline strategies in success rate (\texttt{Accuracy}), and efficiency (\texttt{\#token}). Compared to the best baseline (Reflexion), Data-Copilot achieved an +11.2\% improvement in accuracy and a -75\% reduction in token consumption. Besides, as shown in the last two rows of \Cref{main_result_multi_agent}, Data-Copilot can seamlessly integrate with agent strategies such as ReAct and Reflexion into the workflow deployment. For example, when combined with ReAct, Data-Copilot invokes an interface, obtains an intermediate result, and then reasons to invoke the next interface, improving its performance by +1.3.



\textbf{Data-Copilot Reduces Repetitive Generation in the Real World.} In the real world, most requests are similar or even repetitive. As shown in~\Cref{main_result_multi_agent}, Data-Copilot can save 70\% of token consumption since its output only contains interface names and arguments. In contrast, baseline strategies have to repetitively generate complete code for each request, consuming many more tokens.



\subsection{Why Data-Copilot Brings Improvements?}

\textbf{Data-Copilot exhibits superior performance in complex scenarios.} We categorize test requests by entity count: \texttt{single entity, multiple entities, and multiple entities with complex relations} (statistics in~\Cref{Three Types of Test set}). As \Cref{table:structure} shows, with single entities, improvement is minimal (4\%). However, with multiple entities and complex relations, improvements reach 5.3\% and 19\%, respectively. Baselines often generate \textbf{logically incorrect code} and \textbf{omit critical steps}, while Data-Copilot simply invokes versatile interfaces, reducing real-time response complexity through pre-designed interfaces.



\begin{table}[t!]   
    \centering
    \small
    \setlength\tabcolsep{2pt} 
    \renewcommand\arraystretch{1.1}
    \begin{tabular}{ll|cc} 
    \toprule[1pt]
    & & GPT-3.5 & GPT-4 \\ \cline{3-4}
    \multicolumn{2}{l|}{\textbf{Direct Code Generation}}& 28.3 & 70 \\  \hline
    & \multicolumn{3}{c}{\textbf{\makecell[l]{Workflow Deployment LLM}}} \\ 
    & & GPT-3.5 & GPT-4 \\ \cline{2-4}
    \multirow{2}{*}{\textbf{\makecell[l]{Interface Design LLM}}} & GPT-3.5 & 31.9 & 49.6 \\
    & GPT-4 &70.2 & 74.6 \\ 
    \bottomrule[1pt]
    \end{tabular}
    \caption{We explore different LLM combinations for interface design and workflow deployment.}
    \hspace{0cm}
    \vspace{-0.5cm}
    \label{different LLM for two stages}
\end{table}

\textbf{The Quality of Interface is the Key Factor.} We analyze the effects of using different LLM combinations for two stages: interface design and workflow deployment. As shown in~\Cref{different LLM for two stages}, we observe when using a weaker LLM for interface design, the effectiveness is significantly diminished. For example, using \texttt{GPT-3.5-turbo} (1st stage) and \texttt{GPT-4-turbo} (2nd stage) resulted in a score of only 49.6, which is even lower than directly using GPT-4-turbo for direct code generation (70.2). This phenomenon can also be seen in other combinations. After manually checking, we find interfaces designed by GPT-3.5-turbo are prone to failure and exhibit poor generalizability. Therefore, we chose to use \texttt{GPT-4-turbo} for interface design and optimization, while employing \texttt{GPT-3.5-turbo} for real-time deployment, achieving a balance between accuracy and efficiency.

\begin{table}[t!]
    \centering
    \small
    \setlength\tabcolsep{4pt} 
    \begin{tabular}{lc}
        \toprule[1pt]
        \textbf{Methods}  & \textbf{\texttt{Accuracy}}(\%) \\
        \midrule
        Direct Code Generation     & 28.5     \\ 
        Data-Copilot            & 70.2     \\ \hline
        $w/o$ Self-Exploration Request            & 35.9   \\ 
        $w/o$ Interface Optimization         & 42.1   \\ 
        $w/o$ Interface-Code Hybrid         & 63.8   \\ 
        \bottomrule[1pt]
    \end{tabular} 
    \caption{We ablate three modules from Data-Copilot.}\label{ablation_study_table}
    \hspace{0cm}
    \vspace{-0.5cm}
\end{table}

\section{Conclusion}

We propose Data-Copilot, a code-centric data analysis agent. It generates code for large-scale data processing and creates interface modules through data exploration, improving real-time request success. It autonomously designs universal interfaces for various data types and invokes them for reliable problem-solving. Experiments show higher success rates with lower token consumption. 
\\

\section*{Limitations} 
Data-Copilot proposes a new paradigm for addressing the data-related task, through LLM. But we want to highlight that it still remains some limitations or improvement spaces:

1) \textbf{Online Design Interface.} The essence of Data-Copilot lies in effective interface design, a process that directly affects the effectiveness of subsequent interface deployments. Currently, this interface design process is conducted offline. Therefore, it is crucial to explore how to design the interface online and deploy it simultaneously. It greatly broadens the application scenarios of Data-Copilot.

2) \textbf{System stability} The interface deployment process can occasionally be unstable. The main source of this instability is because LLM is not fully controllable. Despite their proficiency in generating the text, LLMs occasionally fail to follow the instructions or provide incorrect answers, thus causing anomalies in the interface workflow. Consequently, finding methods to minimize these uncertainties during the interface dispatch process should be a key consideration in the future.

3) \textbf{Data-Copilot possesses the potential to handle data from other domains effectively.} Currently, our focus is on developing an LLM-based agent for the data domain. Due to the limited access to data, we chose the China Financial Data, which includes stocks, futures, finance, macroeconomics, and financial news. Although these data all belong to the financial domain, the data volume is extremely large, and the data schemas are highly different. The corresponding user's requests are also diverse, which poses a great challenge to the current LLM. However, Data-Copilot has adeptly accomplished this task. Therefore, we believe Data-Copilot also possesses the potential to effectively handle data from other domains.






\bibliography{custom}

\newpage
\appendix



\renewcommand\thefigure{\Alph{section}\arabic{figure}}    
\setcounter{figure}{0}    
\renewcommand\thetable{\Alph{section}\arabic{table}}    
\setcounter{table}{0}   
\noindent\textbf{Appendix}

\section{Preprocessing of Workflow Deployment}\label{experimetns details}




\textbf{Intent Analysis} To accurately comprehend user requests, Data-Copilot first parses the time, location, data object, and output format of user requests, which are critical to data-related tasks. For example, the request is: \texttt{"I want to compare the GDP and CPI trend in our area over the past five years"}, Data-Copilot parses it as: \texttt{"Draw a line chart of China's national GDP and CPI per quarter from May 2019 to May 2024 for comparison"}. To achieve this, we first invoke an external API to obtain the local time and network IP address, then feed this supportive information into LLMs along with the original request to generate the parsed result.  

\begin{figure*}[t] 
    \centering
    \includegraphics[width=0.95\textwidth]{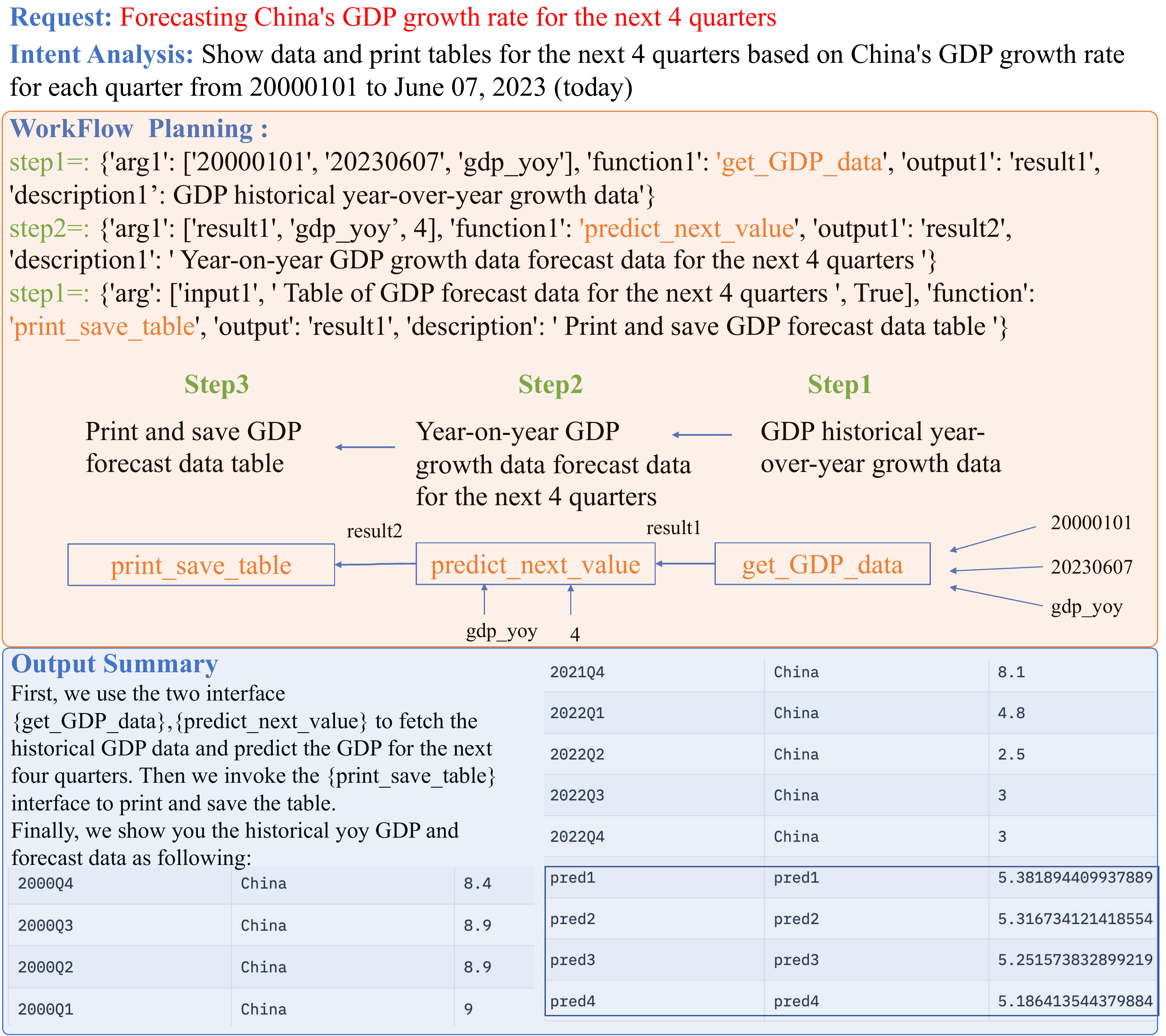}
    \caption{Data-Copilot deploys workflows to solve users' prediction request. It invokes three interfaces step by step and generates arguments for each interface.}
    \label{fig:fig_case_pred}
\end{figure*}

\textbf{Multi-form Output} Upon execution of the workflow, Data-Copilot yields the desired results in the form of graphics, tables, and descriptive text. Additionally, it also provides a comprehensive summary of the entire workflow. It greatly enhances the interpretability of the whole process, as the interface workflow is easy for humans to read and inspect. As the example shown in Figure~\ref{fig:fig_case_pred}, the request is \texttt{"Forecasting China's GDP growth rate..."}. Data-Copilot first interprets the user's intent. Then it deploys a three-step workflow: 1) Invoking \texttt{get-GDP-data()} interface to acquire historical GDP data. 2) Invoking \texttt{predict-next-value()} interface for forecasting. 3) Visualizing the output.

\section{Experiments Details}\label{huamn detail}
\subsection{Baselines and Experiments Details}\label{Baselines details}

We compare Data-Copilot with \ding{173} direct code generation (Direct-Code) and various agent strategies: \ding{174} ReAct~\citep{yao2022react}: LLMs iteratively combine reasoning and code execution. \ding{175} Reflexion~\citep{shinn2023reflexion}: LLMs refine responses based on compiler feedback, limited to two iterations. \ding{176} Multi-agent collaboration~\citep{wu2023autogen, hong2023metagpt, liang2023encouraging}: Three LLM agents (two coders, one manager) collaborate to generate solutions. Data-Copilot uses the same LLM as all baselines when invoking pre-designed interfaces. Detailed prompts are in \Cref{prompt baselines}.

\subsection{The Definitions of Three Complexity Levels for Testset}\label{Three Types of Test set} 
We categorize our test set into three types based on the number of entities involved: single entity, multiple entities, and multiple entities with complex relations (e.g., loop calculations). The ststics of three subset are shown in~\Cref{four tasks and three complexities}.
\begin{itemize}
    \item Single entity: Requests involve a single entity and can be resolved step-by-step. E.g., "query based on a specific condition". 
    \item Multiple entities: Requests require processing multiple entities simultaneously. E.g., "compare a certain metric across multiple entities". 
    \item Multiple entities with complex relations: Involving multiple entities and containing loops, nesting, and other intricate logic. E.g., "List the top 10 stocks by yesterday's price increase that also in the internet industry."
\end{itemize}

\begin{table*}[h!]
    \centering  \small
    \caption{Statistics of four task types and three complexity levels on our test set.} \label{four tasks and three complexities}
    \begin{tabular}{c|l|cccc}
    \toprule[1pt]
    \textbf{\multirow{6}*{\rotatebox{270}{Task Types}}} & \multirow{3}*{\makecell[l]{}} & \multicolumn{4}{c}{\textbf{Request Complexity}} \\
    \cline{3-6}
    & & \makecell[c]{Single Entity} & \makecell[c]{Multiple Entities} & \makecell[c]{Multi-entities with Complex Relation} & \makecell[c]{Overall} \\
    \cline{3-6}
    & \makecell[l]{Stock} & 79 & 106 & 87 & 272 \\
    & \makecell[l]{Fund} & 55 & 36 & 22 & 113 \\
    & \makecell[l]{Corporation} & 97 & 30 & 4 & 131 \\
    & \makecell[l]{Other} & 6 & 12 & 13 & 31 \\
    & \makecell[l]{Total} & 237 & 184 & 126 & 547 \\
    \bottomrule[1pt]
    \end{tabular}
    \hspace{0cm} 
\end{table*}

\begin{table}[t!]
    \centering
    \small
    \setlength\tabcolsep{3pt} 
    \caption{Human evaluation on the human-proposed requests and synthesized requests across four dimensions.}
    \begin{tabular}{lcccc}
        \toprule[1pt]
         & \textbf{\makecell[l]{Task\\Difficulty}} & \textbf{\makecell[l]{Request\\Rationality}} & \textbf{\makecell[l]{Expression\\Ambiguity}} & \textbf{\makecell[l]{Answer\\Accuracy}} \\
        \midrule
        Seed Set   & 3.5  & 4.3 &4.4 & -  \\ 
        Test Set   & 3.9  & 4.2 &4.0 &4.8   \\ 
        \bottomrule[1pt]
    \end{tabular}
    \hspace{0cm}
    \label{human evaluation on seed set and testset}
\end{table}

\subsection{The Quality of Self-exploration Requests.} \label{quality of request}
To assess the quality of self-exploration requests, we invited four additional graduate students to manually evaluate our test set (546 requests) and human-proposed requests (173 seed set) on four criteria: \texttt{task difficulty, request rationality, expression ambiguity, and answer accuracy}. We provide detailed guidance for each criterion in \Cref{guidance of human evaluation}. The results of two sets are shown in~\Cref{human evaluation on seed set and testset}. We observed that the synthesized requests exhibit a comparable quality to human-proposed requests, with slightly higher difficulty, ambiguity, and similar rationality. It ensures that our test set can reflect most of the real-world demands. Besides, the evaluators gave a high score of 4.8 on the label (answer table) of our test set, which also ensures the accuracy of our dataset.

\subsection{The detail of GPT-4o Evaluation}\label{detail of GPT-4o Evaluation} 
\begin{table}[h!]
    \centering
    \small
    \setlength\tabcolsep{5pt} 
    \caption{Comparison of Human and GPT-4o Evaluations}
    \begin{tabular}{lccc}
        \toprule[1pt]
         & \textbf{} & \textbf{\makecell[l]{Human\\Evaluation}} & \textbf{\makecell[l]{GPT-4o\\Evaluation}} \\
        \midrule
        Average Score & & 65.8 & 67.2 \\ 
        Correlation Coefficient & & 0.894 & 0.894 \\ 
        N(score $\geq$ 60) & & 35 cases & 38 cases \\ 
        \bottomrule[1pt]
    \end{tabular}
    \hspace{0cm}
    \vspace{-0.5cm}
    \label{tab:human_gpt_evaluation_comparison}
\end{table}

As described in~\Cref{Experiments Settings}, the evaluation contains three aspects: \textit{Data-table}, \textit{Image}, and \textit{Efficiency}.

\textbf{Data-table Evaluation}: For each request, we use GPT-4o to compare predicted data table with human-annotated table. If GPT-4o identifies any inconsistencies, the judgment is False. For example,  
a request: \textcolor{blue}{\textit{``Please show me the China’s GDP ....''}} with its labeled data-table: \textit{[2023: 17.8 , 2022: 17.9 , 2021: 17.8 , \textcolor{red}{2020: 14.6}]}. Predicted data: \textit{[2023: 17.8 , 2022: 17.9 , 2021: 17.8]}. Based on them, judgment of GPT-4o is \textbf{\textit{False}}.

\textbf{Image Evaluation}: We design a comprehensive evaluation checklist for GPT-4o-based image scoring, comprising 5 main categories with 10 sub-dimensions. It includes (1) numerical points, (2) lines, (3) axes, (4) aesthetics of image layout, and (5) chart design, e.g., \textit{Chart type, Color usage, Proportion and scale, Labels and legends, Readability, Completeness, Relevance, Distinctiveness, Data accuracy,} ... If the total score exceeds 60 points, the image is considered to meet expectations (True); otherwise, it is judged as false.

\textbf{Manual Evaluation Vs. GPT-4o Evaluation.} For data-table evaluation, GPT-4o's assessment results are highly accurate as it only needs to compare differences in text modality. For image evaluation, we randomly sampled 50 examples for both human and GPT-4o-based evaluation. As shown in~\Cref{tab:human_gpt_evaluation_comparison}, we calculate the average score, correlation coefficient, and the \#samples >= 60 of the two methods. The results show that the average scores of the two evaluation methods are close (GPT-based: 67.2, Human-based: 65.8), and the two score sequences also have a strong correlation (0.894).

\subsection{Human Evaluation on Test Benchmark} \label{guidance of human evaluation}
We invite four more graduate students to manually evaluate our benchmark according to four criteria: task difficulty, request rationality, expression ambiguity, and answer accuracy. The belief guidance for human evaluation is as follows:
\begin{itemize}
    \item \textbf{Task Difficulty}: The difficulty of the task, whether it requires multiple steps... Scoring Criteria: 5: very difficult,..., 1: easy.
    \item \textbf{Request Rationality}: Whether the request is reasonable, or if it is strange and does not align with human habits... Scoring Criteria: 5: Reasonable, aligns with humans..
    \item \textbf{Expression Ambiguity}: Whether the phrasing of the request is ambiguous, or if the entities involved are unclear... Scoring Criteria: 5: Clear without any ambiguity...
    \item \textbf{Answer Accuracy}: Whether the answer table is correct, detailed, and comprehensive, totally meeting user expectations... Scoring Criteria: 5: Data is totally correct ... 3: Partial...
    
\end{itemize}

\begin{figure}[!t]
    \centering
    \includegraphics[width=0.7\linewidth]{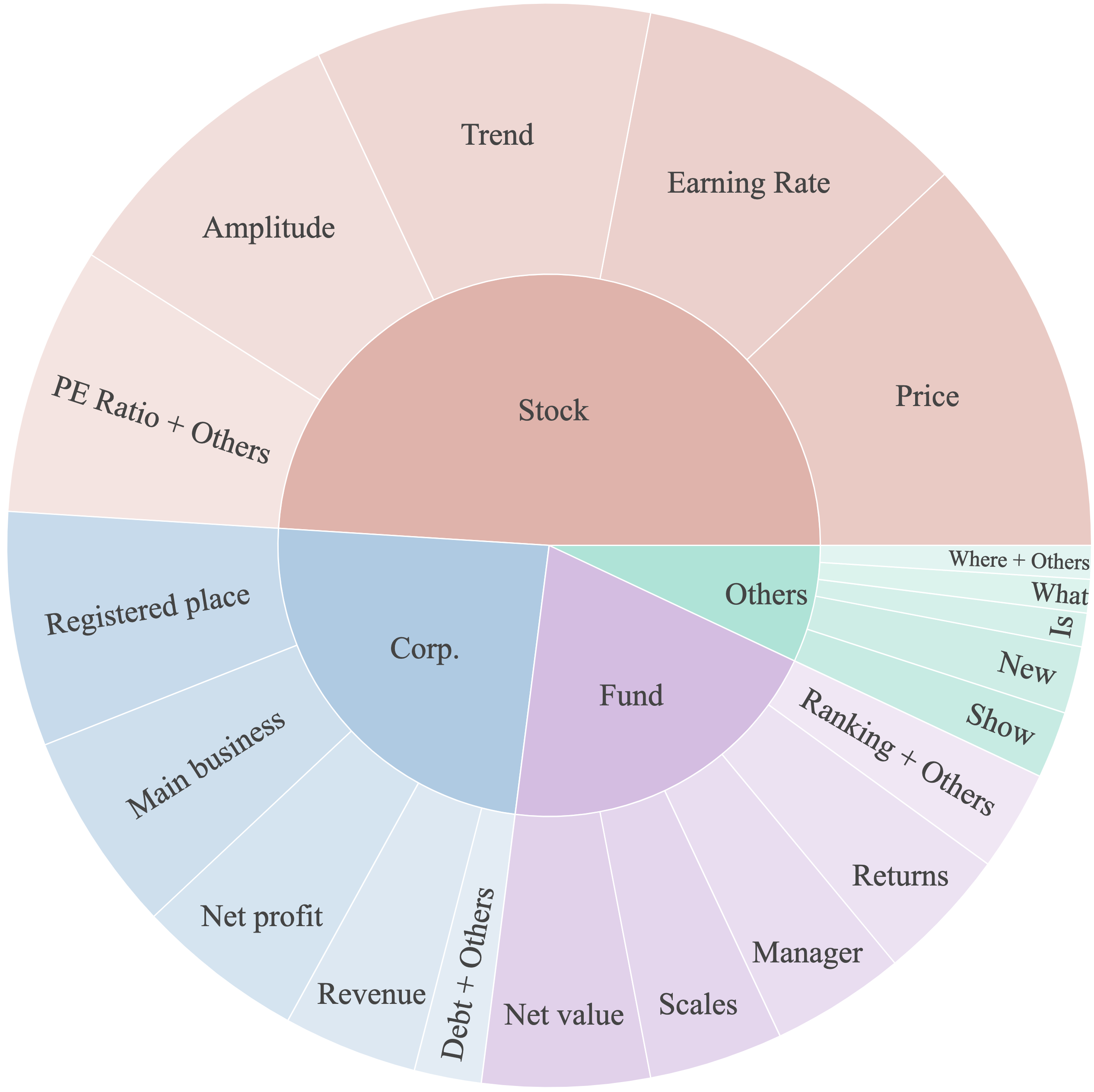}
    \caption{We count the keywords for each request type (Stock, Corp., Fund, Others) in the test set.}
    \label{fig:dataset_pie}
\end{figure}

\subsection{Ablation Study}
As shown in~\Cref{ablation_study_table}, we ablate Data-Copilot from three aspects: \ding{172} \textbf{We ablate the self-exploration request process}. Instead, we directly used seed requests for interface design and optimization. It leads to a 34.3 performance drop. We observe that too few seed requests are insufficient to design universe interfaces. \ding{173} \textbf{We ablate the interface optimization}. It means every successfully designed interface was retained in the interface library. We observe that performance is also significantly affected (-28 points). Without interface optimization, there are many similar interfaces, which hurt the effect of workflow invocation. \ding{174} \textbf{Interface-Code Hybrid Generation}: Data-Copilot can only invoke interfaces during workflow deployment. Without a hybrid generation manner, it shows a 6.4-point performance drop, which is caused by ``uncovered'' requests. It highlights our flexibility in addressing different types of requests.

\subsection{Expanding to Other Programming Languages}
In addition to the Python language, Data-Copilot exhibits excellent scalability, allowing for easy switching to other programming languages by simply regenerating the corresponding interfaces. We tested three programming languages: Python, C++, and Matlab. The results indicate that Python performed the best (\texttt{Accuracy}: 70.2), followed by C++ (54.2), with Matlab (36.5) yielding the poorest results. Upon examination, we found that Matlab code often suffers from formatting errors or dimensional discrepancies in the data, rendering the program non-executable. We speculate this may be related to a lack of sufficient Matlab code in the pre-training corpus. So ultimately, we opted for Python for Data-Copilot.

\section{Related Works}
In the recent past, breakthroughs in large language models (LLMs) such as GPT-3, GPT-4, PaLM, and LLaMa~\citep{brown2020language,chowdhery2022palm, zhang2022opt, Zeng2023GLM,touvron2023llama, Long2022InstructGPT, gpt4,  wei2022emergent,zhang2024tablellm} have revolutionized the field of natural language processing (NLP). These models have showcased remarkable competencies in handling zero-shot and few-shot tasks along with complex tasks like mathematical and commonsense reasoning. The impressive capabilities of these LLMs can be attributed to their extensive training corpus, intensive computation, and alignment mechanism~\citep{Long2022InstructGPT,wang2022supernaturalinstructions, wang2022selfinstruct}.

\textbf{LLM-based Agent} Recent studies have begun to explore the synergy between external tools and large language models (LLMs). Tool-enhanced studies~\citep{schick2023toolformer, Gao2022PALPL, qin2023tool, Hao2023ToolkenGPTAF, qin2023toolllm,hou2023large} integrate external tools into LLM, thus augmenting the capability of LLMs to employ external tools. Several researchers have extended the scope of LLMs to include the other modality~\citep{wu2023visual, surís2023vipergpt, Shen2023HuggingGPTSA, liang2023taskmatrixai, huang2023audiogpt}. In addition, there are many LLM-based agent applications~\citep{xie2023openagents}, such as CAMEL~\citep{li2023camel}, AutoGPT\footnote{\url{https://github.com/Significant-Gravitas/Auto-GPT}}, AgentGPT\footnote{\url{https://github.com/reworkd/AgentGPT}}, BabyAGI\footnote{\url{https://github.com/yoheinakajima/babyagi}}, BMTools\footnote{\url{https://github.com/OpenBMB/BMTools}}, LangChain\footnote{\url{https://github.com/hwchase17/langchain}}, Agentverse~\citep{chen2023agentverse}, Autoagent~\citep{Chen2023AutoAgentsAF}, MetaGPT~\citep{hong2023metagpt}, AutoGEN~\citep{wu2023autogen}, etc. Most of them are focused on daily tools or code generation and do not consider the specificity of data-related tasks. Except for learning to operate the tools, several contemporaneous studies \citep{cai2023large, qian2023creator} have proposed to empower LLMs to create new tools for specific scenarios like mathematical solving and reasoning. These impressive studies have revealed the great potential of LLM to handle specialized domain tasks.

\textbf{Applying LLM To Data Science} Apart from these studies, the application of large models in the field of data science has garnered significant interest among researchers~\citep{maddigan2023chat2vis,valverde2024advanced,gu2024data,liu2024llms,10670425,zhang2023large,ahn2024data,inala2024data,liu2024ai,xie2024waitgpt,wu2024daco,guo2024investigating,cao2024spider2,lu2024large,ye2024dataframe,sui2024table,ford2024charting,chen2024viseval,weng2024datalab,shen2024ask}. FLAME~\citep{joshi2023flame} investigates the feasibility of using NLP methods to manipulate Excel sheets. StructGPT~\citep{jiang-etal-2023-structgpt} explore reasoning abilities of LLM over structured data. LiDA~\citep{dibia2023lida} and GPT4-Analyst~\citep{cheng2023gpt,ma2023demonstration} focus on automated data exploration. Besides, many reseraches~\citep{liu2023comprehensive, chang2023prompt, dong2023c3,almheiri2024data}, like Sheet-Copilot~\citep{li2023sheetcopilot}, BIRD~\citep{li2024can}, DAIL-SQL~\citep{gao2023text}, DIN-SQL~\citep{pourreza2023din}, PET-SQL~\citep{PET-SQL}, DB-Copilot~\citep{wang2023dbcopilot}, MAC-SQL~\citep{wang2023mac}, ACT-SQL~\citep{zhang2023act}, ChatBI~\citep{ChatBI}, CodeS~\citep{CodeS}, SQLPrompt~\citep{sqlprompt}, ChatDB~\citep{hu2023chatdb}, SQL-PaLM~\citep{sun2023sql}, and DB-GPT~\citep{xue2023dbgpt}, EHRAgent~\citep{EHRAgent} apply LLMs to Text2SQL and table rasoning. Chain-of-Table~\citep{wang2024chain} proposes a step-by-step reasoning strategy based on the table. Some researchers also focus on designing various benchmarks and evaluation methods~\citep{lai2023ds,zhang2024benchmarking,sahu2024insightbench,yang2024matplotagent,ma2024spreadsheetbench,gu2024blade,jing2024dsbench,hu2024infiagent} for LLMs in data science.

\section{Visualization} \label{visualization}

We provide several cases in this section to visualize workflow deployed by Data-Copilot, which includes queries about diverse sources (stocks, company finance, funds, etc.) using different structures (parallel, serial, and loop Structure).

\begin{figure*}[!htp] 
    \centering
    \includegraphics[width=0.8\textwidth]{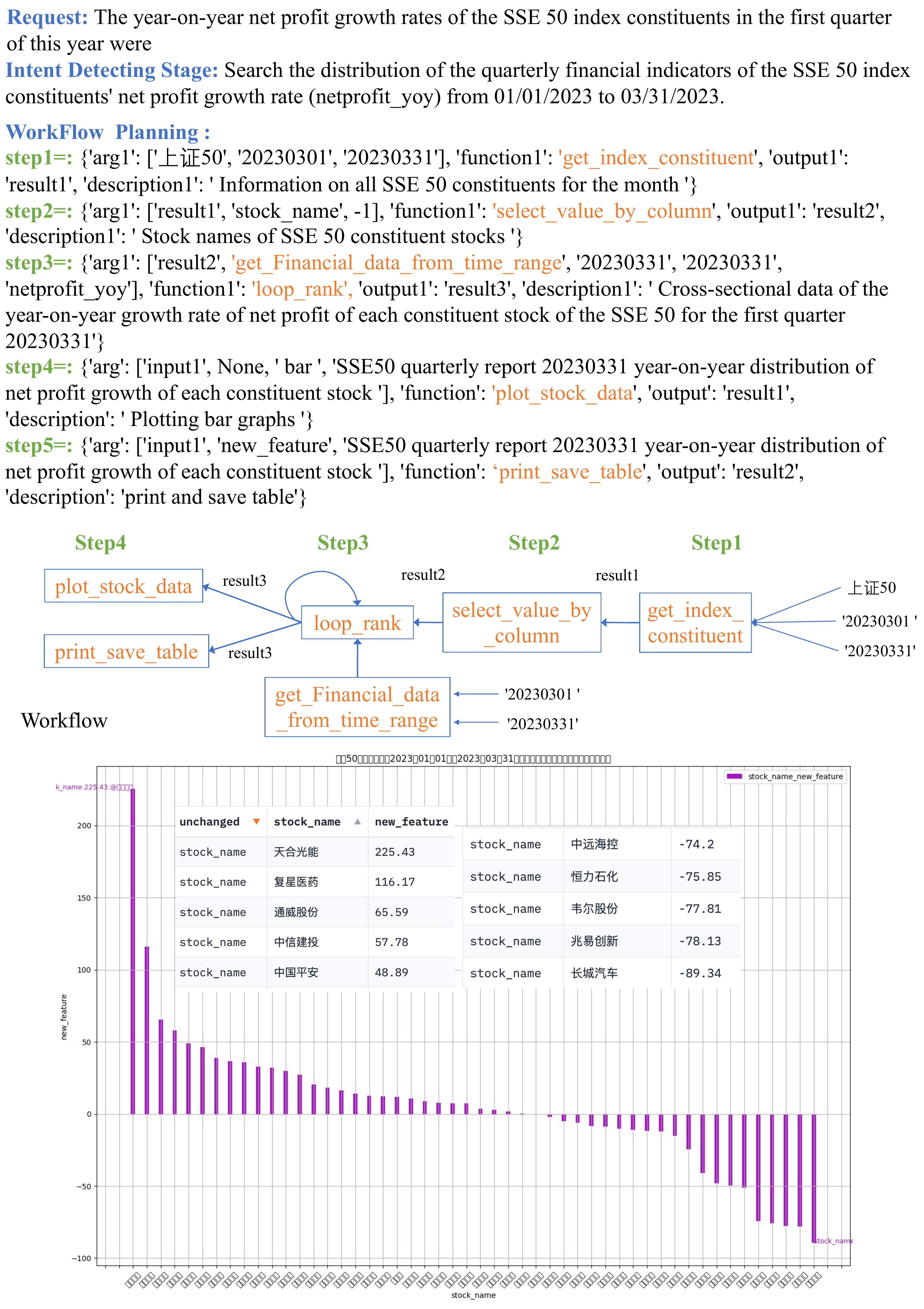}
    \caption{For complex requests about stock financial data, Data-Copilot deploys a loop workflow to solve user requests and finally outputs images and tables in parallel.}
    \label{fig:fig_case_rank}
\end{figure*}

\begin{figure*}[!htp] 
    \centering
    \includegraphics[width=0.8\textwidth]{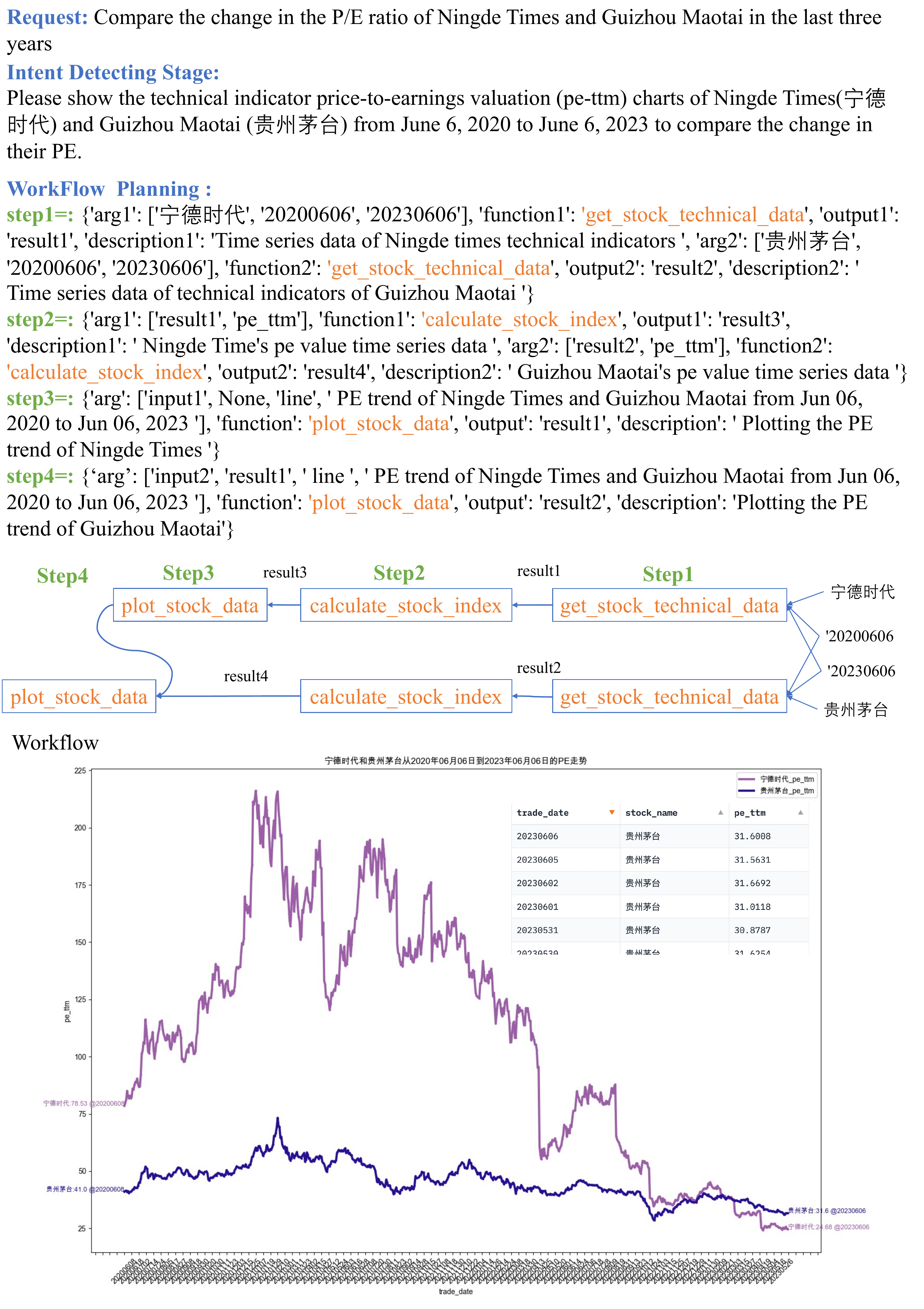}
    \caption{Data-Copilot acquires the P/E data of two stocks in parallel and plots them in a graph.}
    \label{fig:fig_case_stock}
\end{figure*}

\textbf{Different Structures}
As shown in Figure~\ref{fig:fig_case_rank},\ref{fig:fig_case_stock}, Data-Copilot deploys different structural workflows based on user requirements. In Figure~\ref{fig:fig_case_rank}, the user proposes a complex request, and Data-Copilot deploys a loop structure to implement the financial data query of each stock, and finally outputs a graph and table in parallel. In Figure~\ref{fig:fig_case_stock}, Data-Copilot proposes a parallel structure workflow to meet user request (the demand for comparison in user request) and finally draws two stock indicators on the same canvas. These concise workflows can cope with such complex requests well, which suggests that the data exploration and workflow deployment process of Data-Copilot are rational and effective.

\begin{figure*}[!htp] 
    \centering
    \includegraphics[width=0.9\textwidth]{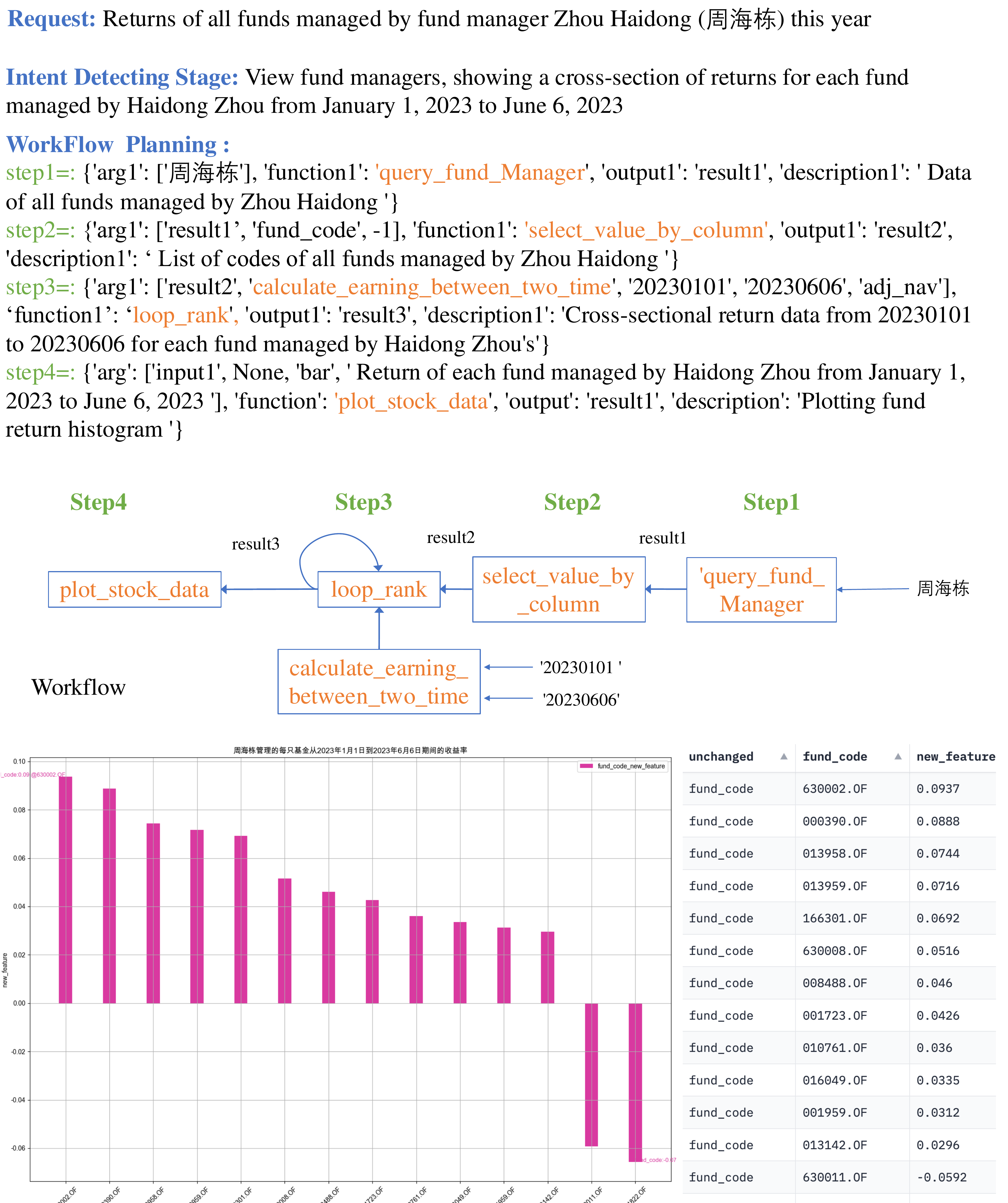}
    \caption{Data-Copilot also has access to fund data and can query the returns of all funds managed by the fund manager.}
    \label{fig:fig_case_fund}
\end{figure*}

\begin{figure*}[!htp] 
    \centering
    \includegraphics[width=0.8\textwidth]{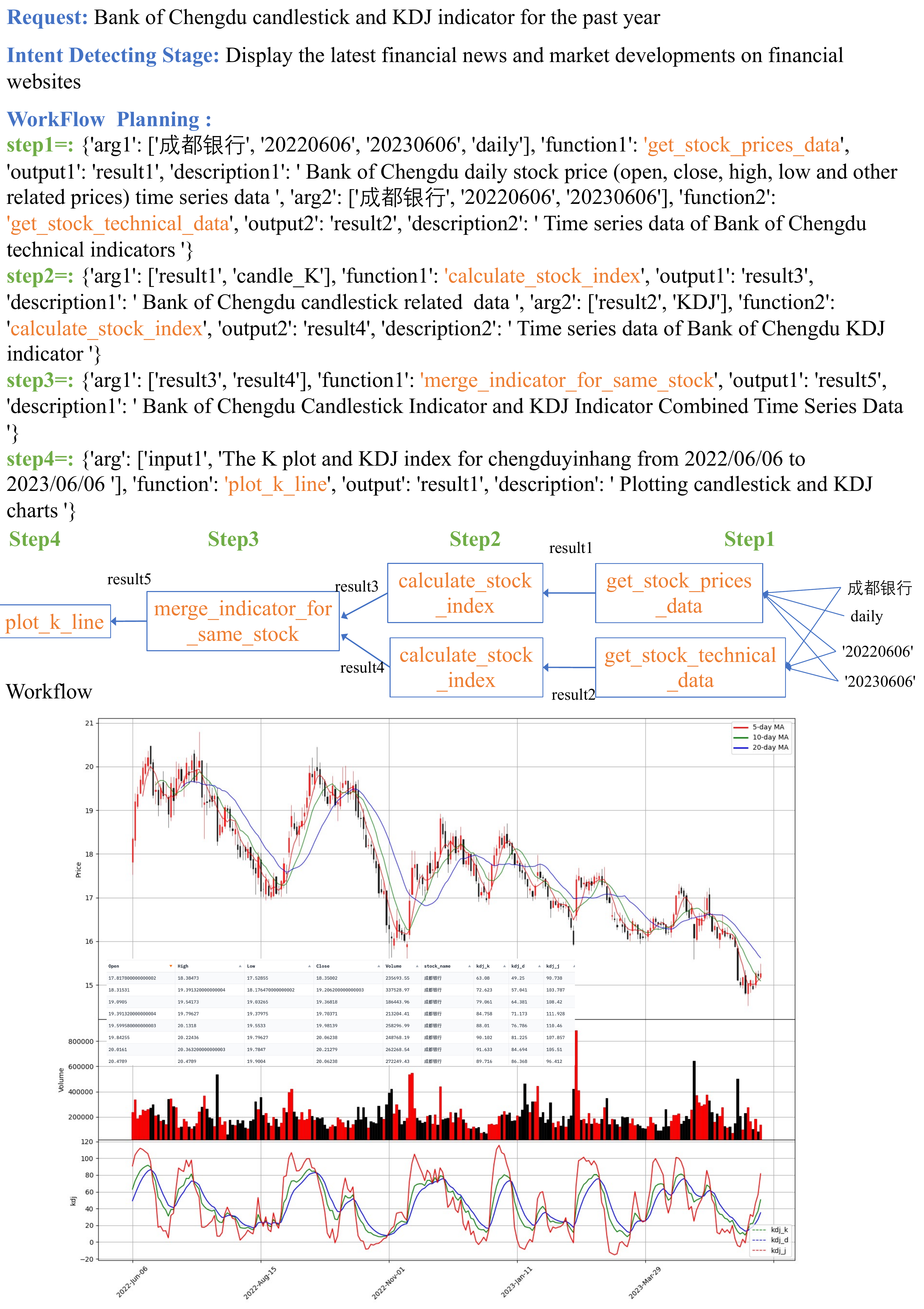}
    \caption{Data-Copilot can plot multiple indicators in a single graph by deploying workflows in parallel.}
    \label{fig:fig_case_kline}
\end{figure*}

\begin{figure*}[!htp] 
    \centering
    \includegraphics[width=0.8\textwidth]{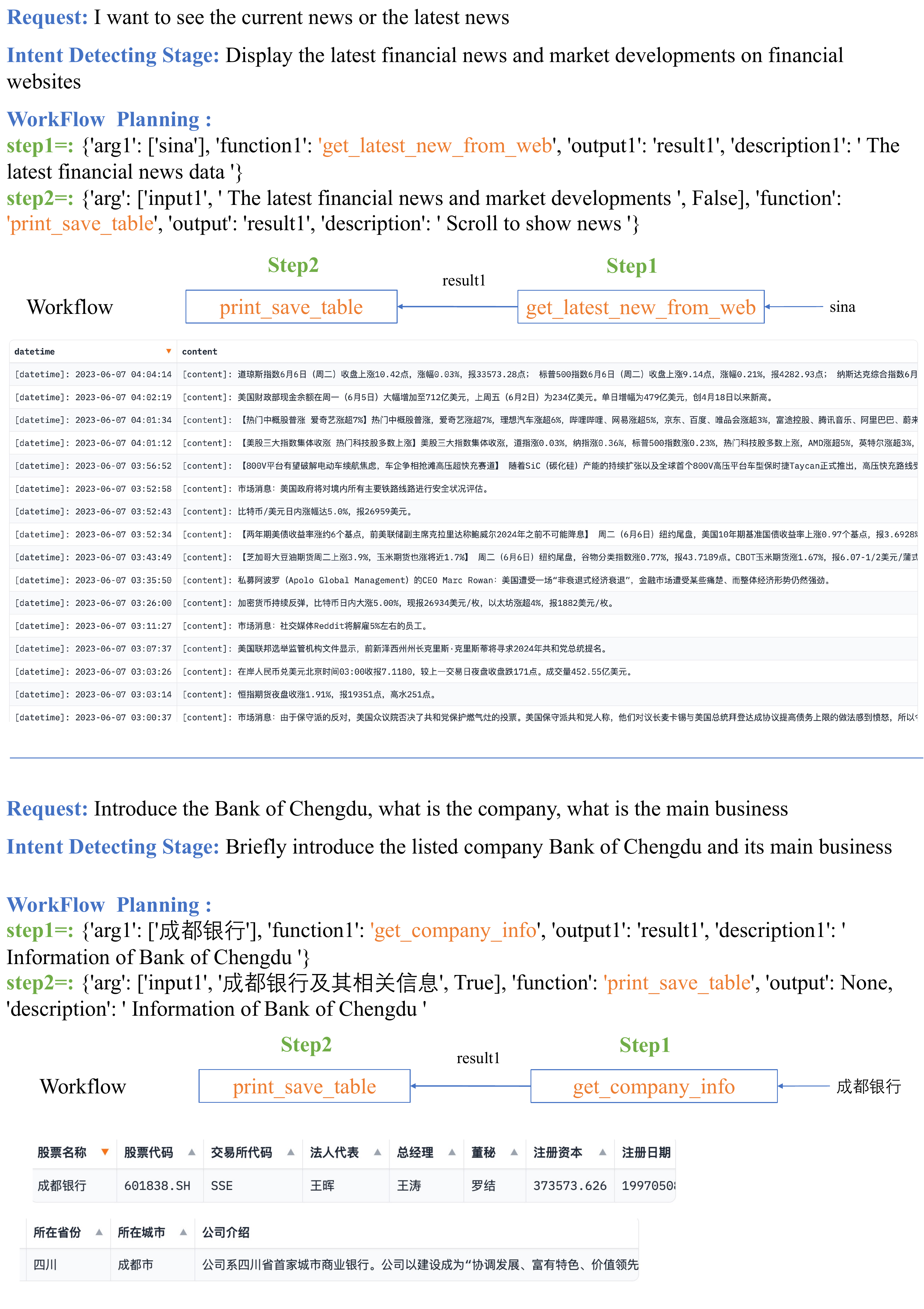}
    \caption{Data-Copilot can provide the latest financial news and company information by deploying the corresponding workflows.}
    \label{fig:fig_case_livenew}
\end{figure*}

\textbf{Diverse Sources}
Figure~\ref{fig:fig_case_fund}, \ref{fig:fig_case_kline}, \ref{fig:fig_case_livenew} demonstrate that Data-Copilot is capable of handling a large number of data sources, including stocks, funds, news, financial data, etc. Although the formats and access methods of these data types are quite different, our system efficiently manages and displays the data through its self-designed versatile interface, requiring minimal human intervention.

\section{Detailed Prompts}
We provide detailed prompts for our Data-Copilot and baselines. Specifically, for the data exploration, we provide detailed prompts in~\Cref{prompt for design} for four phases: self-exploration, interface design, and interface optimization. For the workflow deployment phase, we also provide prompts for three key procedures in~\Cref{prompt for Dispatch}: Intent Analysis, Task Selection, and Planning Workflow. Additionally, we outline prompts for all baselines in~\Cref{prompt baselines}: Direct-Code, ReAct, Reflection, and Multi-Agent Collaboration strategies.

\lstset{
    framesep = 20pt,
    rulesep = 10pt,
    backgroundcolor = \color[RGB]{245,245,244},
    breaklines = true,
    breakindent = 0pt,
    basicstyle = \ttfamily\small,
    escapeinside = {(*@}{@*)} 
}
\begin{figure*} 
\subsection{Algorithm Flow For Interface Design and Optimization} \label{design_Algorithm}
\begin{lstlisting} 
(*@\textbf{Step 1: Interface Design and Testing}@*)

1. (*@\textbf{One-to-One Interface Design}@*)
   * Design interfaces one-by-one using synthesized requests (request 1----> interface 1)
   * Generate a series of initial interfaces with single functionalities
2. (*@\textbf{Test Case Generation}@*)
   * For each designed interface:
      > Sample a sub data-table from relevant data sources
      > Generate test requests based on data schemas and sampled sub-tables
      > For example: For an interface: "GDP_retrieves(Year, Country)"
         - Sample a data point from the GDP table: China, 2023, $13 trillion
         - Generate test request: "What is China's GDP in 2023?"
         - Use the sampled sub-table as the expected answer
      > Repeat this process K times for each interface, generating K test cases
3. (*@\textbf{Interface Evaluation}@*)
   * Use the generated test cases to evaluate the usability of each interface
   * Record interfaces that pass the tests and their testing results

(*@\textbf{Step 2: Interface Optimization and Compiler Feedback Evaluation}@*)


1. (*@\textbf{Similar Interface Retrieval}@*)
   * Analyze the functionality and parameters of all interfaces
   * Identify pairs of interfaces with similar or overlapping functionalities
2. (*@\textbf{Interface Merging Decision}@*)
   * For each pair of similar interfaces, assess the necessity and feasibility of merging
   * Consider functional coverage, usage scenarios, and complexity of the interfaces
3. (*@\textbf{Interface Merging Execution}@*)
   * Design a new merged interface ensuring:
      > The new interface covers all functionalities of the original two interfaces
4. (*@\textbf{Merged Interface Evaluation}@*)
   * Evaluate the merged interface using test cases generated in the previous steps
   * Collect compiler feedback on the merged interface
   * Verify:
      > Whether the new interface can correctly handle all test cases of the original interfaces
      > Whether the output is consistent with the original interfaces in both format and content
5. (*@\textbf{Feedback-Based Self-Optimization}@*)
   * If the merged interface fails evaluation:
      > Guide the LLMs to analyze compiler feedback
      > Perform self-reflection based on the feedback and two old interfaces
      > Optimize the interface design to address issues
   * Repeat the evaluation-reflection-optimization cycle multiple times
   * Abandon the merging attempt after multiple fails 
\end{lstlisting}
\end{figure*}

\subsection{Example for Data Exploration}\label{example for data exploration}
\lstset{
    framesep = 20pt,
    rulesep = 10pt,
    backgroundcolor = \color[RGB]{245,245,244},
    breaklines = true,
    breakindent = 0pt,
    basicstyle = \ttfamily\small,
    escapeinside = {(*@}{@*)} 
}
\begin{lstlisting}(*@\textbf{Explore Data by Self-Exploration Phase}@*)
###Instruction: Given some data and its description, please mimic these seed requests and generate more requests. The requests you generate should be as diverse as possible, covering more data types and common needs.
###Seed Request: {(*@\color{blue}{request1, request2,...}@*)}
###Parsing file for GDP_Data:{
    Description: (*@\color{blue}{This data records China's annual and quarterly GDP...}@*),
    Access Method: (*@\color{blue}{\emph{pro.cn-gdp(start-time, end-time, frequency,...)}}@*), 
    Output Schema: (*@\color{blue}{Return 9 columns, including \emph{quarter}, \emph{gdp}, \emph{gdp-yoy}...}@*),
    Usage:{ 
        Example: (*@\color{blue}{\emph{pro.cn-gdp(start-q='2018Q1', end-q='2019Q3',..)}}@*),
        First Row: {(*@\color{blue}{2019Q4, 990, ..}@*)}, Last Rows: {(*@\color{blue}{2018Q4, 900, ..}@*)}}}
###Parsing file for Stock_data:{...}, .... 
\end{lstlisting}

\begin{figure*} 
\subsection{Prompts For Data Exploration} \label{prompt for design}
\begin{lstlisting} 
(*@\textbf{Explore Data by self-exploration Phase}@*)
###Instruction: Given some data and its description, please mimic these seed requests and generate more requests. The requests you generate should be as diverse as possible, covering more data types and common needs.

###Seed Request: {(*@\color{blue}{request1, request2,...}@*)}
###parsing file for GDP-Data:{
    Description: (*@\color{blue}{This data records China's annual and quarterly GDP...}@*),
    Access Method: (*@\color{blue}{You can access the data by \emph{pro.cn-gdp(start-time, end-time, frequency,...)}, start-time means....}@*), 
    Output Schema: (*@\color{blue}{The data return 9 columns, including \emph{quarter}: quarter, \emph{gdp}: cumulative GDP, \emph{gdp-yoy}: quarterly Year-on-Year growth rate, \emph{pi},...}@*),
    Usage:{ 
            Example: (*@\color{blue}{\emph{pro.cn-gdp(start-q='2018Q1', end-q='2019Q3', frequency='quarter')}}@*),
            First Row: {(*@\color{blue}{2019Q4  990865.1    6.10, ...}@*)},
            Last Rows: {(*@\color{blue}{2018Q4  900309.5    6.60, ...}@*)}}}
###parsing file for stock_data: {...}


(*@\textbf{Interface Design}@*): 
###Instruction: You are an experienced program coder. Given a request and some existing interfaces, you should use these interfaces to solve the request or design new interfaces to resolve my request. 

(1) You should define the name, function, inputs, and outputs of the interface. Please describe the functionality of the interface as accurately as possible and write complete implementation code in the new interface.
(2) Finally please explain how to resolve my request using your newly designed interfaces or existing interfaces in the neural language. 

###Output Format:
Your newly designed interfaces:
Interface1={Interface Name: {(*@\color{blue}{name}@*)}, 
            Function description: {(*@\color{blue}{This interface is to ...}@*)}, 
            Input: {(*@\color{blue}{argument1: type, argument2: type, ...}@*)}, 
            Output: {(*@\color{blue}{pd.DataFrame}@*)} }
Interface2=....

The solving process for request: {(*@\color{blue}{To fullfil this request, I design a interface ...}@*)}
###The user request: {(*@\color{blue}{input request}@*)}
###Data Files: {(*@\color{blue}{All data parsing files}@*)}
###The existing interfaces: {(*@\color{blue}{all interfaces in library}@*)}

(*@\textbf{Interface Optimization}@*)
###Instruction: Please check that the interface you have designed can be merged with any existing interfaces in the library. 
(1) You should merge interfaces with similar functionality and similar input and output formats into a new interface. 
(2) You can use parameters to control the different inputs, if you want to merge two interfaces.
(3) Please explain your reason for merging and output all interfaces in the library after merging. 
(4) If you don't think a merge is necessary, then just add new interfaces into the existing interface library and output them all.

###New interfaces: {(*@\color{blue}{interfaces}@*)}
###Existing interfaces: {(*@\color{blue}{interface1, interface2, interface3, ..}@*)}
###Output Format:
The reasons for merging: {(*@\color{blue}{reason}@*)}
Interfaces after merging: {(*@\color{blue}{interface1, interface2, optimized interface3, ..}@*)}
\end{lstlisting}
\end{figure*}

\begin{figure*} 
\subsection{Prompts For Workflow Deployment} \label{prompt for Dispatch}
\begin{lstlisting} 
(*@\textbf{Intent Analysis Phase}@*)
### Analysis Prompt: Please parse the input request for time, place, object, and output format. You should rewrite the instruction according to today's date. The rewritten new instruction must be semantically consistent and contain a specific time and specific indicators. 

### Output Format: Rewritten Request. (Time:%s, Location:%s, Object:%s, Format:%s).

### User Request: Today is {(*@\color{blue}{Timestamp}@*)}. The user request is {(*@\color{blue}{Input Request}@*)}.
Please output a Rewritten Request.


(*@\textbf{Task Selection}@*)
###Select Prompt: Please select the most suitable task according to the given Request and generate its task_instruction in the format of task={task_name: task_instruction}. There are four types of optional tasks. [fund_task]: used to extract and process tasks about all public funds. [stock_task]: for extracting and processing tasks about all stock prices, index information, company financials, etc., [economic_task]: for extracting and processing tasks about all Chinese macroeconomic and monetary policies, as well as querying companies and northbound funds, [visualization_task]: for drawing one or more K-line charts, trend charts, or outputting statistical results. 

###Output Format: task1={%s: %s}, task2={%s: %s}

###User Request: {(*@\color{blue}{Rewritten Request}@*)}.
Please output a task plan for this request.

(*@\textbf{Planning Workflow}@*)
###Planning prompt: Please use the given interface (function) to complete the Instruction step by step. At each step you can only choose one or more interfaces from the following interface library without dependencies, and generate the corresponding arguments for the interface, the arguments format should be strictly in accordance with the interface description. The interface in the later steps can use results generated by previous interfaces.

###Output Format:
Please generate as json format for each step:step1={"arg1": [arg1,arg2...], "function1": "%s", "output1": "%s", "description1": "%s"}, step2={"arg1": [arg1,arg2..], "function1": "%s", "output1": "%s", "description1": "%s"}, ending with ###.

###User Request: {(*@\color{blue}{Task Instruction}@*)}.
Please output an interface invocation for this instruction.
\end{lstlisting}
\end{figure*}

\begin{figure*} 
\subsection{Prompts For Baselines}\label{prompt baselines}
\begin{lstlisting}
(*@\textbf{Direct-Code LLM}@*)
###Instruction: You are an artificial intelligence assistant. Given some data access methods and a user request, you should write a complete Python code to fulfill the user's request. Your code must completely fulfill all the user's requirements without syntax errors!

###User Request: {(*@\color{blue}{User request}@*)}
###Data files: {(*@\color{blue}{All data files}@*)}
Please solve the request by Python Code.

(*@\textbf{Step-by-Step ReAct}@*)
###Instruction: You are an artificial intelligence assistant. Given some data access methods and a user request, please think step by step and generate your thoughts and actions for each step, and then finally realize the user's request.
###User Request: {(*@\color{blue}{User request}@*)}
###Data files: {(*@\color{blue}{All data files}@*)}

###Thought Prompt: Please think about the next action that should be taken to handle the user request.
### {(*@\color{blue}{Thought: I need to ...}@*)}
###Action Prompt: Based on your previous thoughts, please generate a complete Python code to accomplish what you just planned.
### {(*@\color{blue}{Action: def get-data()....@*)}
###Observation Prompt: Please summarize the results of the code execution just now and think about whether this result accomplishes what you planned for this step.
### {(*@\color{blue}{Action: Yes, I observed that this function successfully fetched the data...@*)}
....


(*@\textbf{Step-by-Step Reflexion}@*)
###Instruction: You are an artificial intelligence assistant. Given some data access methods and a user request, please think step by step and then generate your thoughts and actions for each step. After the execution of your current action, you need to reflect on the results until your current plan has been successfully completed. Then you think about the next step and then generate your next action, and finally realize the user's request.
###User Request: {(*@\color{blue}{User request}@*)}
###Data files: {(*@\color{blue}{All data files}@*)}

###Thought Prompt: Please think about the next action that should be taken to handle the user request.
### {(*@\color{blue}{Thought: I need to ...}@*)}
###Action Prompt: Based on your previous thoughts, please generate a complete Python code to accomplish what you just planned.
### {(*@\color{blue}{Action: def get-data()....@*)}
###Observation Prompt: Please record the compiler's return results just now and think about whether this result accomplishes what you planned for this step.
### {(*@\color{blue}{Action: No, I observe that the compiler returns an error...@*)}
### Reflection Prompt: Please reflect on the error returned by the compiler and regenerate a new Python code to resolve the issue. If the compilation passes without any errors, reflect on whether the current result is what you planned to do.
### {(*@\color{blue}{Action: I revise my solution as follows: def get-data2()....@*)}
....
\end{lstlisting}
\end{figure*}

\begin{figure*} 
\begin{lstlisting}
(*@\textbf{Multi-agent collaboration}@*)
###Instruction For Manager: You are the manager of the project team and you need to lead your team to fulfill user requests. You have two experienced programmers under you (Programmer-A and -B) and you need to assign them the same or different tasks according to the user request, then organize the discussion, and finally solve the problem.
###Instruction For Agent1/Agent2: You are an experienced programmer. Your team has a colleague who is also a programmer and a manager. You need to write code according to the manager's arrangement, discuss it with them, improve your program, and get a consensus conclusion.
###User Request: {(*@\color{blue}{User request}@*)}
###Data files: {(*@\color{blue}{All data files}@*)}


--------  Phase1 Discussion for Task Assignment -------
###Task Assignment Prompt For 3 Agents: Now let's start discussing how to solve user problems. The speaking order is manager, A and B. After a round of discussion, the manager decides on a preliminary solution and assigns specific tasks to the two programmers
###Manager Response: {(*@\color{blue}{To fulfill the request... @*)}
###Programmer A Response: {(*@\color{blue}{Yes, I agree with the manager. We should first,....@*)}
###Programmer B Response: {(*@\color{blue}{That's a good solution. We also should...@*)}
###Manager Response: {(*@\color{blue}{After discussion, Programmer A should..., B should ....@*)}


--------  Phase2 Coding and Discussion-------
###Prompts For 3 Agents: Two programmers complete their respective tasks as assigned by their manager. Then the manager organizes a discussion on whether the two options and their running results are reasonable. Lastly, two programmers need to regenerate the code.
###Programmer A Response: {(*@\color{blue}{Solution1 Generated by A@*)}
###Programmer B Response: {(*@\color{blue}{Solution1 Generated by B@*)}
###Prompts For 3 Agents: Please evaluate two solutions. {(*@\color{blue}{Solution1, Result1, Solution2, Result2@*)}
###Programmer A Evaluation: {(*@\color{blue}{I think ...@*)}
###Programmer B Evaluation: {(*@\color{blue}{I agree with ...@*)}
###Programmer A Regeneration: {(*@\color{blue}{Revised Solution1@*)}
###Programmer B Regeneration: {(*@\color{blue}{Revised Solution2@*)}

--------  Phase3 Discussion-------
....

--------  Phase4 Final Decision-------
###Prompts For manager: Please output a complete and correct solving code for user request based on the previous discussions of the group.
###Manager Decision: {(*@\color{blue}{Final Solution@*)}
\end{lstlisting}
\end{figure*}

\section{Case Study}
Detailed cases of data exploration and workflow deployment are presented in \Cref{case interface design,case interface dispatch}.
\begin{figure*} 
\subsection{Cases for Data Exploration} \label{case interface design}
\begin{lstlisting}
---- (*@\textbf{The First Request: Interface Design}@*) -----
Input Request: Get the daily opening price data of Bank of Hangzhou from Jan 23, 2018 to Mar 13, 2019 and plot the trend.

Response: First, we need to design a function to get the opening price data from the bank's database. Then, we need another function to process this data and prepare them for graphical representation. Finally, we need a function to plot the trend chart. 

New Design Interfaces:
{(*@\color{blue}{Interface1={Name:getBankData(), Function:.., Input:.., Output:..}@*)}
{(*@\color{blue}{Interface2={Name:prepareDataForPlotting(), Function:.., Input:.., Output:..}@*)}
{(*@\color{blue}{Interface3={Name:plotData(), Function:.., Input:.., Output:..}@*)}

The solving process for request: These three new interface functions will meet your needs. In your program, you would first get the data using getBankData, then process the data using prepareDataForPlotting, and finally use plotData to plot the trend.
Complete code: {(*@\color{blue}Complete code using three new interfaces...@*)}

---- (*@\textbf{The Second Request: Interface Definition Phase}@*)-------
Input Request: Get the monthly closing price data of Ningde Times from January 23, 2018 to March 13, 2019, and plot the trend.

Response: To fulfill this requirement, we design a new interface function. In addition, I need to use two existing interfaces to fulfill your requirement.

New Designed Interfaces:
{(*@\color{red}{Interface4={Name: getStockData()..}@*)}

Existing Interfaces
{(*@\color{blue}{Interface1={Name:getBankData(), ...}@*)}
{(*@\color{blue}{Interface2={Name:prepareDataForPlotting(), ...}@*)}
{(*@\color{blue}{Interface3={Name:plotData(), ...}@*)}

The solving process for request: You need to call these functions to meet your needs. First, use new interface {(*@\color{red}{getStockData()}@*)} to get the data of Ningde Times, then use existing interface {(*@\color{blue}{prepareDataForPlotting()}@*)} to process this data, and finally use {(*@\color{blue}{plotData()}@*)} to plot the trend...

Complete code: {(*@\color{blue}Complete code using three new interfaces...@*)}

---- (*@\textbf{Interface Optimization Phase}@*) -------
The reason for merging: Yes, these interface functions can be merged and reformulated to eliminate duplicate functions. We can merge (*@\color{blue}{getBankData()}@*) and (*@\color{blue}{getStockData()}@*) into a more generic function (*@\color{red}{get-financial-data()}@*), which can fetch any financial data, including bank and stock data. In addition, we can keep the prepareDataForPlotting and plotData functions, as they can also be applied to different types of financial data]

{(*@\color{purple}{Interface1={Interface Name:get-financial-data(), ...}@*)}
{(*@\color{blue}{Interface2={Interface Name:prepareDataForPlotting(), ...}@*)}
{(*@\color{blue}{Interface3={Interface Name:plotData(), ...}@*)}
\end{lstlisting}
\end{figure*}

\begin{figure*} 
\subsection{Cases for Real-time Request} \label{case interface dispatch}
\begin{lstlisting}
### Request: Today is 2019-03-13. The user request is to please help me plot the stock price trend of Guizhou Maotai from January 23, 2018 to today.

### Rewritten Request: Please help me plot the closing price of Guizhou Maotai from Jan 23, 2018 to Mar 13, 2019 (Time:20180123-20190313, Location: China, Object: Guizhou Maotai stock price, Format: Line).

###Plan: task1={"stock_task": "Get the time series data of Guizhou Maotai's stock price from 20180123 to 20190313"}, task2={"visualization_task": "Plot the stock price of Guizhou Maotai from 20180123 to 20190313 on a line chart"}

###Function Call: 
            step1={
                "arg1": ["Guizhou Maotai", "20180123", "20190313", "daily"], 
                "function1": "get_stock_prices_data", 
                "output1": "result1", 
                "description1": "Guizhou Maotai daily stock price (open, close, high, low, etc.) time series data"}, 
            step2={
                "arg1": ["result1","close"], 
                "function1": "calculate_stock_index", 
                "output1": "result2v, 
                "description1": "Guizhou Maotai's closing price time series data"}, 
            step3={
                "arg": ["result2", null, "line", "Guizhou Maotai closing price chart from January 23, 2018 to March 13, 2019"], "function": "plot_stock_data", 
                "output": "result1", 
                "description": "Guizhou Maotai Stock Price Line Chart"
            } ###


\end{lstlisting}
\end{figure*}

\begin{figure*}[!htp] 
    \centering
    \includegraphics[width=0.75\textwidth]{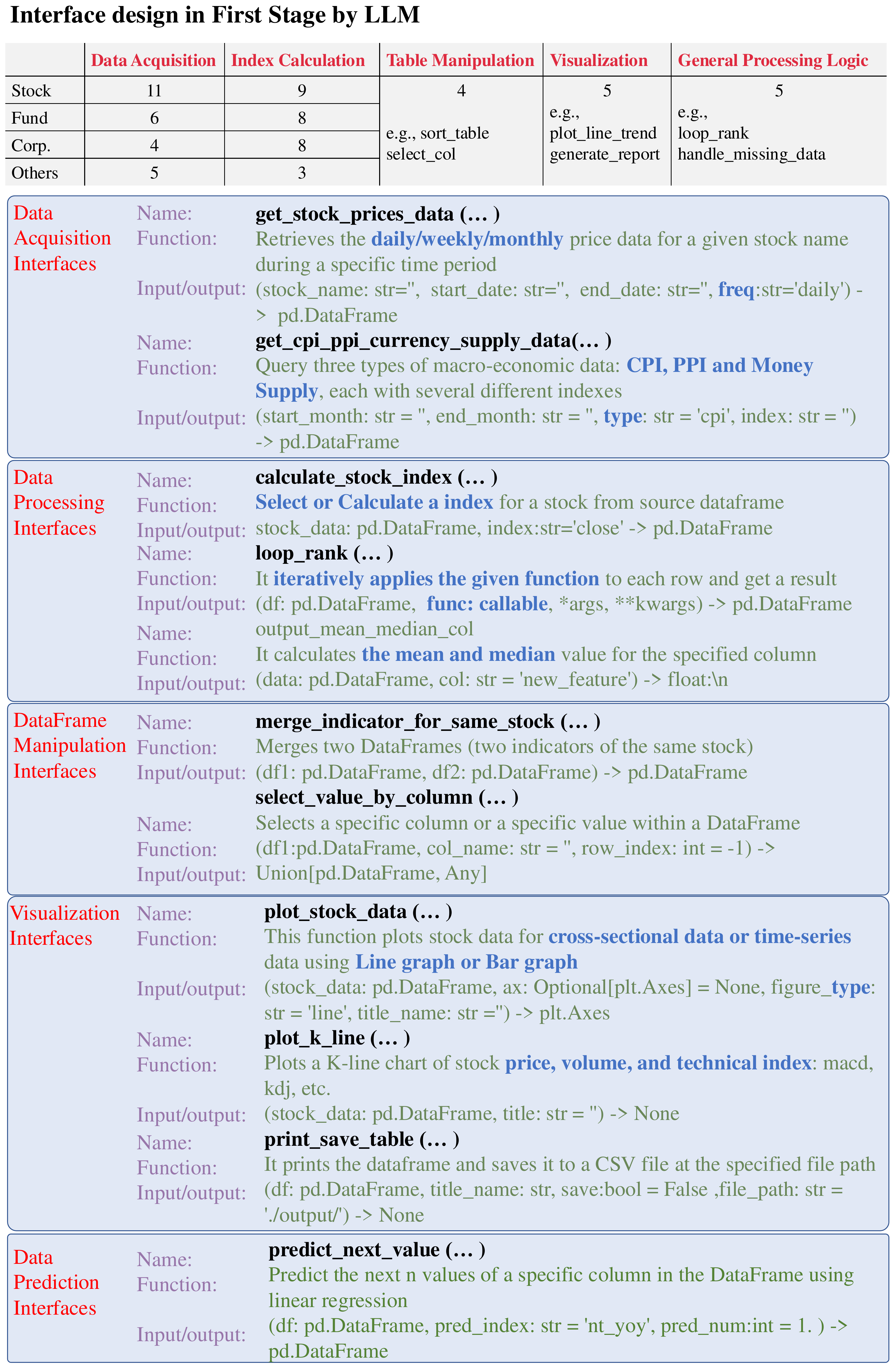}
    \caption{We have listed some of Data-Copilot's self-design interfaces, including five categories. Most of the interfaces are a combination of several simple interfaces. For example, the text marked in blue indicates that the interface includes multiple functions.}
    \label{fig:interface}
\end{figure*}

\begin{figure*}[!htp] 
    \centering
    \includegraphics[width=0.7\textwidth]{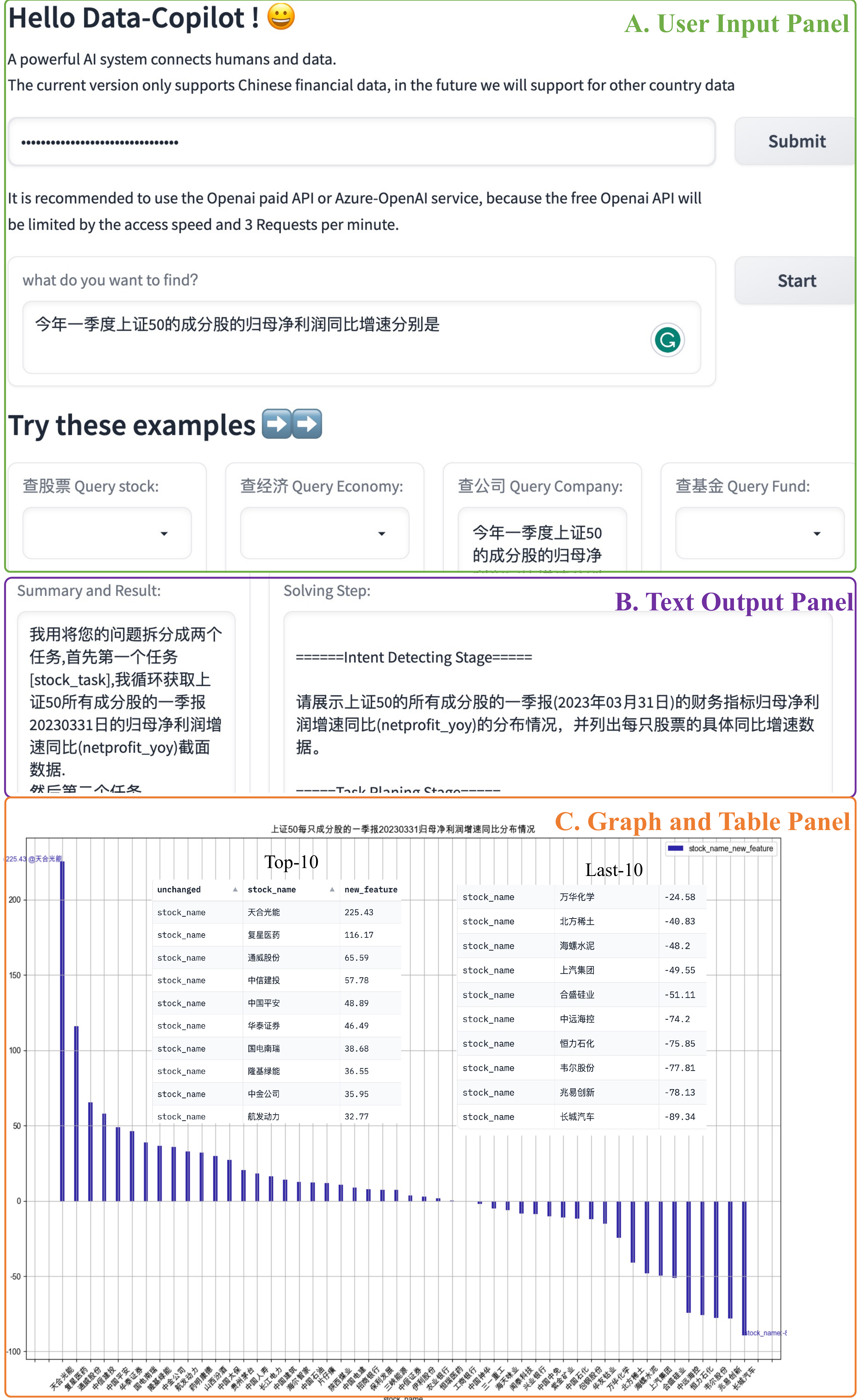}
    \caption{The user interface of our system. The green box (A) is the user input panel, and the purple (B) and red parts (C) are the results returned by the system.}
    \label{d4}
    \label{fig:3}
\end{figure*}

\end{document}